\PassOptionsToPackage{table}{xcolor}

\documentclass[preprint,12pt]{elsarticle}
\usepackage{dirtytalk}
\usepackage[utf8]{inputenc}
\usepackage{xcolor}
\usepackage{rotating}
\usepackage{amssymb}
\usepackage{amsmath,amssymb,amsfonts}
\usepackage{algorithmic}
\usepackage{graphicx}
\usepackage{textcomp}
\usepackage{longtable}
\usepackage{hyperref}
\usepackage{geometry}
\usepackage{array}
\usepackage{tabularx}
\usepackage{comment} 
\usepackage{multirow}
\usepackage{pdflscape}
\usepackage{subcaption}
\usepackage{colortbl}
\usepackage{arydshln} 
\usepackage{tcolorbox}
\usepackage{algorithm}
\usepackage{float}

\journal{Journal of Systems and Software}
\begin{document}
\begin{frontmatter}
\title{FAIREDU: A Multiple Regression-Based Method for Enhancing Fairness in Machine Learning Models for Educational Applications}
\author[inst1,inst2]{Nga Pham}
\affiliation[inst1]{organization={Faculty of Information Technology},%Department and Organization
            addressline={Dainam University}, 
            city={Hanoi},
            %postcode={00000}, 
            %state={State One},
            country={Vietnam}}
\author[inst3]{Minh Kha Do}
\author[inst1,inst2]{Tran Vu Dai}
\author[inst2]{Pham Ngoc Hung}
\author[inst4]{Anh Nguyen-Duc}
\affiliation[inst2]{organization={Faculty of Information Technology},%Department and Organization
            addressline={VNU University of Engineering and Technology}, 
            city={Hanoi},
            %postcode={22222}, 
            %state={State Two},
            country={Vietnam}}
\affiliation[inst3]{organization={School of Computing, Engineering and Mathematical Sciences},%Department and Organization
            addressline={Latrobe University}, 
            %city={Bø I Telemark},
            %postcode={22222}, 
            %state={State Two},
            country={Australia}
            }
\affiliation[inst4]{organization={Faculty of Information Technology},%Department and Organization
            addressline={University of South Eastern Norway}, 
            city={Bø I Telemark},
            %postcode={22222}, 
            %state={State Two},
            country={Norway}}
\begin{abstract}
%% Text of abstract
Fairness in artificial intelligence and machine learning (AI/ML) models is becoming critically important, especially as decisions made by these systems impact diverse groups. In education, a vital sector for all countries, the widespread application of AI/ML systems raises specific concerns regarding fairness. Current research predominantly focuses on fairness for individual sensitive features, which limits the comprehensiveness of fairness assessments. This paper introduces FAIREDU, a novel and effective method designed to improve fairness across multiple sensitive features. Through extensive experiments, we evaluate FAIREDU's effectiveness in enhancing fairness without compromising model performance. The results demonstrate that FAIREDU addresses intersectionality across features such as gender, race, age, and other sensitive features, outperforming state-of-the-art methods with minimal effect on model accuracy. The paper also explores potential future research directions to enhance further the method’s robustness and applicability to various machine-learning models and datasets.
\end{abstract}

\begin{highlights}
    \item FAIREDU addresses fairness across intersectional sensitive features
    \item We explored the characteristics of several education datasets 
    \item We investigated tradeoff among fairness and performance for ML algorithms
\end{highlights}
\end{frontmatter}

\section{Introduction}
\label{sec: Intro}
With the increasing application of Machine Learning (ML) systems across various industries and sectors of society \citep{nguyen-duc_generative_2024}, ensuring the quality of these systems is becoming more important. In the software industry, AI/ML algorithms are potentially transforming how software is developed and operated \citep{peng_fairmask_2023}. As AI/ML takes on a greater role in decision-making processes, particularly with decisions affecting diverse groups, fairness has emerged as a critical concern \citep{b1,peng_fairmask_2023}. Unfair outcomes in AI/ML systems are often viewed as "fairness bugs," and substantial research has been dedicated to detecting and mitigating these biases \citep{b2,b3,b4,b5,b6,b7,b8,b9}. ML algorithms, for example, can introduce biases linked to sensitive features like gender \citep{b11,b12} or race \citep{b9,b11,b13}, disadvantaging historically marginalized groups.

In education, fairness in ML systems extends beyond technical challenges, requiring solutions that address deep-rooted social and structural inequalities \citep{pham_fairness_2024}. Scholars have long studied disparities in educational access and outcomes, particularly focusing on issues like school segregation and achievement gaps \citep{minnaert_1997,engberg_2004,huston_2006,hughes_2013,mahmud_2020}. For instance, it is unfair if students from low-income families consistently score lower due to limited access to resources, or if teacher evaluations and algorithmic grading systems contain biases \citep{pessach_2022,zhai_2020}. Addressing multiple social factors—such as gender, race, socioeconomic status, and disability—is essential for achieving fairness \citep{b28}. However, this is a complex issue, as different subgroups face varying degrees of privilege or disadvantage \citep{b29}. Moreover, there is often a trade-off between fairness and model performance \citep{b8,b32,b33,b34,b35}, and the extent to which current methods balance these two aspects remains unclear, especially when considering multiple sensitive features.

Existing fairness methods fall into three main categories: pre-processing, in-processing, and post-processing \citep{b16,b17}. Pre-processing methods, like Reweighting (RW) \citep{b18} and Disparate Impact Remover (DIR) \citep{b19}, adjust the data before model training. In-processing methods, such as Meta Fair Classifier (META) \citep{b20}, Adversarial Debiasing (ADV) \citep{Mitigating_Unwanted_Biases}, and PR (Prejudice Remover) \citep{b22}, intervene during model training. Post-processing methods, like Equalized Odds Processing (EOP) \citep{b23}, Calibrated Equalized Odds (CEO) \citep{b24}, and ROC (Reject Option Classification) \citep{b25} adjust the model's predictions. Additionally, methods combining multiple stages have been proposed, such as Fair-SMOTE \citep{b4}, MAAT \citep{b15}, and FairMask \citep{b26}. While effective, these methods often focus on a single sensitive feature, which limits their ability to address fairness across intersecting features.

In 2022, Yanhui Li et al. introduced LTDD, a linear-regression-based Training Data Debugging method that enhances fairness by eliminating dependencies between features and sensitive features, making it a simple yet effective solution for real-world applications \citep{b27}. However, LTDD is limited to handling one feature at a time, which can improve fairness for a specific feature while potentially reducing it for others \citep{b17}. Recently, a few studies have focused on fairness for multiple sensitive features. For instance, Zhenpeng Chen et al. proposed a solution to improve fairness by forming sensitive features by combining different sensitive features into subgroups \citep{b17}. Although the combination is quite simple, it provides a solution to address fairness research on a single sensitive feature.

This work proposes a novel method, FAIREDU, that is both simple and effective in addressing fairness across intersectional features within the educational context. Four research questions (RQs) are derived from the research objective:

\begin{enumerate}
    \item RQ1 -  Is there a systematic bias present among sensitive features within educational datasets?
    \item RQ2 - Does the level of fairness vary across different machine learning models?
    \item RQ3 - How does FAIREDU manage multiple sensitive features compared to current state-of-the-art methods?
    \item RQ4 - How effectively does FAIREDU balance fairness and model performance relative to state-of-the-art methods?
\end{enumerate}

FAIREDU works as follows: the method detects the dependency of remaining features on sensitive features based on a multivariate regression model and then removes the dependency to create a new dataset that ensures fairness for all features without reducing model performance. We highlight the key characteristics of our method:
\begin{itemize}
    \item FAIREDU addresses fairness across multiple sensitive features.
    \item FAIREDU handles multiple sensitive features very well.
    \item It applies to both discrete and continuous sensitive features.
\end{itemize}

The rest of this paper is structured as follows. Section 2 presents the background and related work in fairness in ML. Section 3 introduces the FAIREDU method in details. Section 4 describes the experimental setup and methodology used to evaluate FAIREDU. Section 5 presents the research results. Section 6 provides a detailed discussion, where we address the research questions and show the limitations of our approach. Finally, Section 7 concludes the paper and suggests future research directions.
\section{Background}
\subsection{Fairness for Machine Learning Systems}
Fairness has been a topic of extensive philosophical debate for centuries, with no universally accepted definition due to differing perspectives and cultural contexts. As artificial intelligence (AI) and machine learning (ML) systems become increasingly embedded in various aspects of life, they now play a significant role in decision-making processes that directly affect individuals \citep{pessach_2022}. These systems, however, are susceptible to biases, often reflecting the values and prejudices of their human designers. Saxena et al. (2022) note that "fairness in decision-making can be understood as the absence of bias or prejudice against individuals or groups based on inherent characteristics" \citep{b36}. While the precise definition of fairness in AI/ML remains contested, Hutchinson and Mehrabi offer several prominent interpretations that highlight the diversity of thought in this area \citep{b10,b36}. These definitions, summarized in Table \ref{Table Definitions of fairness}, provide a foundation for understanding how fairness is applied in AI/ML systems.

%Requirement Engineering activities
\begin{longtable}{|p{2.5cm}|p{4.5cm}|p{4.5cm}|p{1.5cm}|}
\caption{Definitions of fairness} 
\label{Table Definitions of fairness} \\
\hline
\textbf{Type of Fairness} & \textbf{Definition} & \textbf{Explanation} & \textbf{Ref} \\ 
\hline
\endfirsthead
\multicolumn{4}{c}%
{{\tablename\ \thetable{} -- continued from previous page}} \\
\hline
\textbf{Type of Fairness} & \textbf{Definition} & \textbf{Explanation} & \textbf{Ref} \\ 
\hline
\endhead
\hline
%\multicolumn{4}{|r|}{{Continued on next page}} \\
\hline
\endfoot
\hline
\endlastfoot

Equalized Odds & A predictor \( \hat{Y} \) satisfies equalized odds with respect to protected attribute (sensitive feature) \( A \) and outcome \( Y \), if \( \hat{Y} \) and \( A \) are independent conditional on \( Y \). \( P(\hat{Y}=1|A=0,Y=y) = P(\hat{Y}=1|A=1,Y=y) \), \( y \in \{0,1\} \) & The protected and unprotected groups should have equal rates for true positives and false positives & \citep{b23,b37} \\ 
\hline
Equal Opportunity & A binary predictor \( \hat{Y} \) satisfies equal opportunity with respect to \( A \) and \( Y \) if \( P(\hat{Y}=1|A=0,Y=1) = P(\hat{Y}=1|A=1,Y=1) \) & The protected and unprotected groups should have equal true positive rates & \citep{b23,b37,b39} \\ 
\hline
Demographic Parity & A predictor \( \hat{Y} \) satisfies demographic parity if \( P(\hat{Y}|A=0) = P(\hat{Y}|A=1) \) & The likelihood of a positive outcome should be the same regardless of whether the person is in the protected group (e.g., female) & \citep{b37,b39,b40,b41} \\ 
\hline
Fairness Through Awareness & An algorithm is fair if it gives similar predictions to similar individuals, where& Any two individuals who are similar with respect to a similarity (inverse distance) metric defined for a particular task should receive a similar outcome &  \citep{b37,b39,b40}\\ 
\hline
Fairness Through Unawareness & An algorithm is fair as long as any protected attributes \( A \) are not explicitly used in the decision-making process & & \citep{b37,b39,b41}\\ 
\hline
Treatment Equality & Treatment equality is achieved when the ratio of false negatives and false positives is the same for both protected group categories & & \citep{b37,b43}\\ 
\hline
Test Fairness & A score \( S = S(x) \) is test fair (well-calibrated) if it reflects the same likelihood of recidivism irrespective of the individual’s group membership, \( R \). That is, if for all values of \( s \), \( P(Y=1|S=s,R=b) = P(Y=1|S=s,R=w) \) & For any predicted probability score \( S \), people in both protected and unprotected (female and male) groups must have an equal probability of correctly belonging to the positive class & \citep{b37,b39,b44} \\ 
\hline
Counterfactual Fairness & Predictor \( \hat{Y} \) is counterfactually fair if under any context \( X = x \) and \( A = a \), \( P(\hat{Y}_{(A\leftarrow a)}(U)=y|X=x,A=a) = P(\hat{Y}_{(A\leftarrow a')}(U)=y|X=x,A=a) \) (or all \( y \) and for any value \( a' \) attainable by \( A \)) & Intuition that a decision is fair towards an individual if it is the same in both the actual world and a counterfactual world where the individual belonged to a different demographic group & \citep{b37,b40}\\ 
\hline
Fairness in Relational Domains & fairness criterion that integrates both individual attributes and the relational structures connecting individuals within a specific domain & considering the personal characteristics of each individual alongside the social, organizational, and interpersonal relationships that influence and are influenced by those characteristics & \citep{b37,b45}\\ 
\hline
Conditional Statistical Parity & For a set of legitimate factors \( L \), predictor \( \hat{Y} \) satisfies conditional statistical parity if \( P(\hat{Y}|L=1,A=0) = P(\hat{Y}|L=1,A=1) \) & People in both protected and unprotected (female and male) groups should have an equal probability of being assigned to a positive outcome given a set of legitimate factors \( L \) & \citep{b37,b39,b34}\\ 
\hline

\end{longtable}

\subsection{Sensitive features}
The fairness literature primarily focuses on characteristics of individuals \citep{b47, b5, sweeney_discrimination_2013}. To prevent discrimination during tasks like classification or prediction, certain personal characteristics must be protected; these are known as protected attributes or sensitive features. Common sensitive features include sex, race, age, religion, disability status, and national origin. In real-world applications, ML systems often need to account for multiple sensitive features simultaneously. Based on the values of these sensitive features, individuals can be divided into privileged and unprivileged groups. Typically, the privileged group is associated with favorable labels, while the unprivileged group is more likely to receive unfavorable labels \citep{b17}. The most common sensitive features in the education context are summarized in Table \ref{Table Sensitive}.
\begin{table}[]
%\begin{small}
\begin{center}
\caption{Sensitive features in Education Studies}
\label{Table Sensitive}
%\scalebox{0.8}{ % Giảm kích thước bảng xuống 90%
\begin{tabular}{|p{4cm}|p{10cm}|}
\hline
\textbf{Sensitive feature} & \textbf{Description} \\\hline
\textbf{Gender} & A person's biological sex, which can be male, female, or non-binary\\
\hline
\textbf{Race} & Physical characteristics, such as skin color, hair texture, and facial features, which can be used to categorize people into different racial groups\\
\hline
\textbf{Ethnicity/Disability} & A person's cultural and racial identity can be influenced by factors such as ancestry, language, and shared cultural practices\\
\hline
\textbf{Age} & Person's chronological age\\
\hline
\textbf{Country} & The nation or sovereign state in which a person lives or was born\\
\hline
\textbf{Language} & Person's native language or the language they speak most fluently\\
\hline
\textbf{Income Level} & A person who has a low income or not\\
\hline
\textbf{Year of study (First-gen)} & In education, it is understood as first-year students - subjects who are confused with information about schools and majors \\
\hline
\textbf{Origin} & Place or country of a person's birth or ancestry\\
\hline
\textbf{Parental background} & Parental Education Background refers to the level of formal education that a child's parents or guardians have achieved \\
\hline
\textbf{Home literacy environment} &Home Literacy Environment refers to the availability and quality of reading materials, as well as literacy-related activities and interactions within a child's home\\
\hline
\textbf{Health} & The health status of the learner\\\hline

%\textbf{Voice accent} & Pronunciation of native and non-native speakers will have certain differences that need attention in speech recognition. & \citep{b43} & 1 \\
\end{tabular}
\end{center}
\end{table}
\subsection{Detecting and fixing fairness bugs for AI/ML systems}
Detecting and addressing fairness bugs in AI/ML systems involves a range of strategies, which are broadly categorized into three main approaches: pre-processing, in-processing, and post-processing methods. Pre-processing methods focus on modifying the training data to eliminate biases before the model is trained. Methods in this category include reweighting, resampling, and data transformation to ensure that the dataset does not favor any particular group. In-processing methods integrate fairness considerations directly into the model training process. These methods involve adjusting the learning algorithms to minimize bias, such as through adversarial debiasing, fairness constraints, or incorporating fairness-aware loss functions. Post-processing methods aim to adjust the model’s predictions after training to achieve fair outcomes. This can involve methods like equalized odds processing, and reject option classification, which modify the decision thresholds to ensure fairness across different groups

\textbf{Pre-processing methods:}
\begin{itemize}
   \item RW (Reweighting) \citep{b18} employs differential weighting of training data for each combination of groups and labels to achieve fairness.
    \item DIR (Disparate Impact Remover) \citep{b19} adjusts feature values to enhance fairness while preserving the rank-ordering within groups 
\end{itemize}
    
\textbf{In-processing methods:}
\begin{itemize}
   \item META (Meta Fair Classifier) \citep{b20} employs a meta-algorithm to optimize fairness regarding protected attributes.
   \item ADV (Adversarial Debiasing) \citep{Mitigating_Unwanted_Biases} uses adversarial methods to minimize the presence of protected attributes in predictions, while concurrently maximizing prediction accuracy.
   \item PR (Prejudice Remover) \citep{b22} incorporates discrimination-aware regularization to mitigate the influence of protected attributes.
\end{itemize}
\textbf{Post-processing methods:}
\begin{itemize}
    \item EOP (Equalized Odds Processing) \citep{b23} uses linear programming to calculate probabilities for adjusting output labels, aiming to optimize equalized odds concerning protected attributes.
    \item CEO (Calibrated Equalized Odds) \citep{b24} optimizes the probabilities of modifying output labels based on calibrated classifier score outputs, with the objective of achieving equalized odds.
    \item ROC (Reject Option Classification) \citep{b25} assigns favorable outcomes to unprivileged instances and unfavorable outcomes to privileged instances near the decision boundary, particularly when there is high uncertainty.
\end{itemize}

Additionally, there are three state-of-the-art methods proposed in the SE literature, including Fair-SMOTE \citep{b4}, MAAT \citep{b15}, and FairMask \citep{b26}.
\begin{itemize}
    \item Fair-SMOTE \citep{b4} generates synthetic samples to achieve balanced distributions between different labels and various protected attributes within the training data. Additionally, it removes ambiguous samples from the training set.
    \item MAAT \citep{b15} combines individual models optimized for ML performance and fairness concerning each protected attribute, respectively. It ensures that both fairness and ML performance objectives are met.
    \item FairMask \citep{b26} trains extrapolation models to predict protected attributes based on other data features. Subsequently, it uses these extrapolation models to modify the protected attributes in test data, enabling fairer predictions
\end{itemize}

\section{FAIREDU - A regression-based method for fairness of multiple sensitive features in Education}
\subsection{Idea development}
Assumed that we have an AI/ML model that does classification or produces binary value, denoted as $S_{\text{ML}}$ in Formula \ref{S_ML}, can be defined as a function that maps domain feature vectors $\mathbf{x} = [x_1, x_2, \dots, x_d] \in \mathbb{R}^d$ to class labels $y \in \{0, 1\}$, i.e.,

\begin{equation}
\label{S_ML}
   S_{\text{ML}} : \mathbb{R}^d \rightarrow \{0, 1\}. 
\end{equation}

Typically, for a new input $\mathbf{x}$, $y$ represents the actual label, while $\hat{y} = S_{\text{ML}}(\mathbf{x})$ denotes the label predicted by the ML software.

Building on the effective solution to fairness challenges presented by Li et al. with the LTDD method \citep{b27}, we developed FAIREDU to address scenarios involving multiple sensitive features. Pre-processing methods like LTDD allow for the correction of biases directly within the dataset, ensuring that the data used to train machine learning models is fair and unbiased from the outset. This method is particularly advantageous because it is model-agnostic \citep{ribeiro_anchors_2018}, meaning it can be seamlessly integrated with various types of machine learning algorithms without requiring modifications to the model architecture or training procedures.

In diference to the LTDD method, this approach specifically addresses the intersectionality of sensitive attributes such as gender, race, age, and disability, denoted as \( x_1, \ldots, x_k \), we will use multivariate regression to simultaneously eliminate the dependencies of each non-sensitive feature on all sensitive features. Mathematically, this is as defined in Formula \ref{multiple regression def}

\begin{equation}
    \label{multiple regression def}
           x_i = \beta_0 + \beta_1 \cdot x_1 + \beta_2 \cdot x_2 + \dots + \beta_k \cdot x_k + \epsilon  
\end{equation}

By employing this multivariate regression model, FAIREDU effectively detects and removes the dependencies of the remaining features on all specified sensitive features, thereby enhancing the fairness of machine learning systems in educational contexts. This method ensures a balanced consideration of multiple sensitive features, addressing the complexities introduced by intersectionality and reducing the risk of bias across different groups.

\subsection{Overall architecture of FAIREDU}

\begin{figure}[htbp]
\centerline{\includegraphics[width=1.0\textwidth]{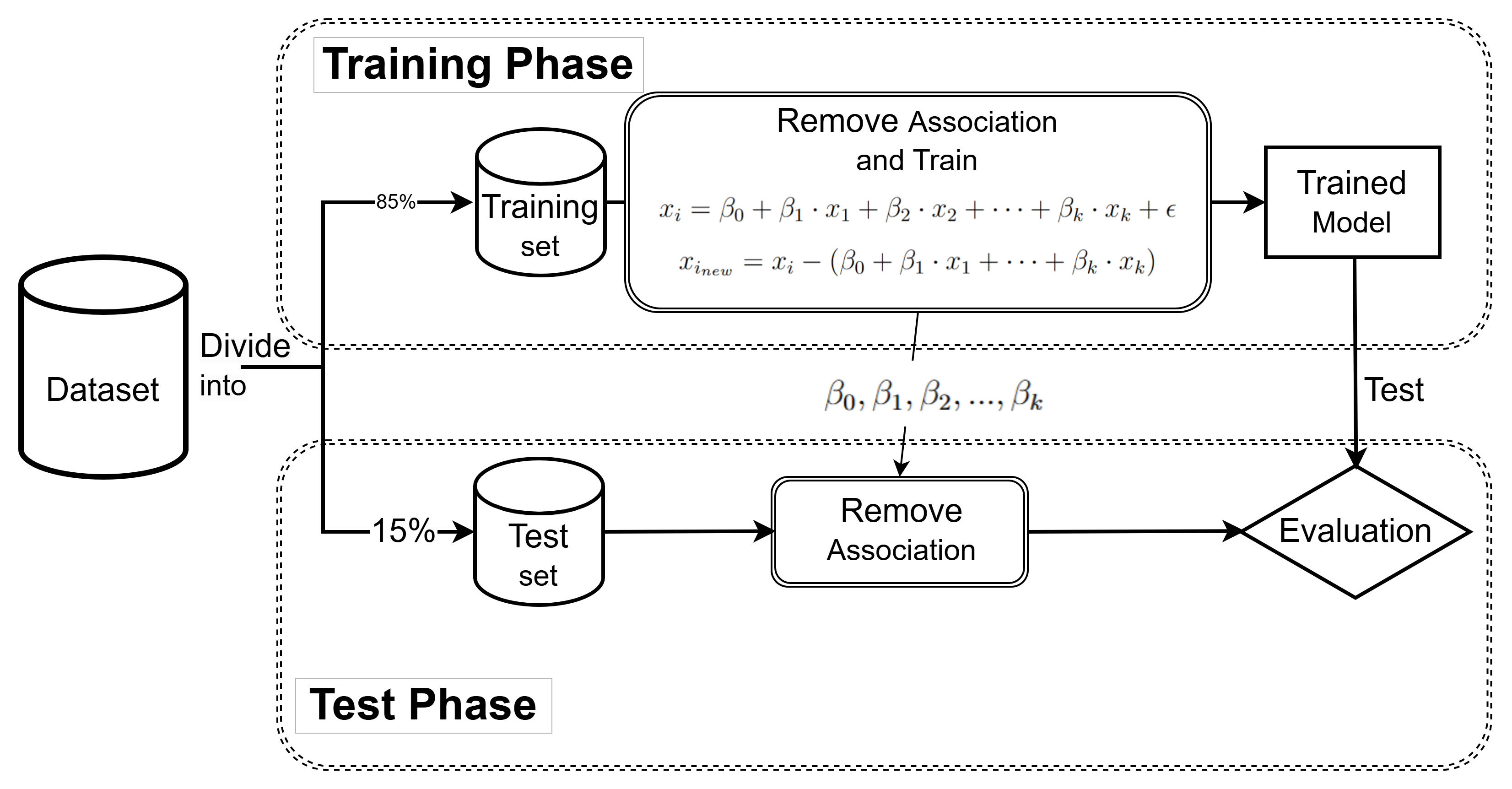}}
\caption{The overall architecture of FAIREDU}
\label{Fig. 1}
\end{figure}

The architecture in Figure \ref{Fig. 1} represents the overall workflow of the FAIREDU model, which is designed to improve fairness in machine learning systems by addressing the dependencies between sensitive features and other features in the dataset. Here's a breakdown of how this figure works in combination with the previously generated explanation:

\begin{enumerate}
    \item Dataset Preparation:
    \begin{itemize}
        \item The process starts with the full dataset, which contains both sensitive and non-sensitive features.
        \item The dataset is divided into two parts:
        \begin{itemize}
            \item \textbf{Training Set (85\%)}: Used for training the model.
            \item \textbf{Test Set (15\%)}: Reserved for testing and evaluating the trained model on unseen data.
        \end{itemize}
    \end{itemize}

    \item Remove Association and Train (Training Set):
     \begin{itemize}
        \item In the training set, the FAIREDU algorithm applies multivariate regression to identify and remove dependencies between non-sensitive features and multiple sensitive features (e.g., gender, race, age).
        \item We assume have k sensitive features $x_1, \ldots, x_k$. For each non-sensitive feature $x_i, k+1 \le i \le d$, we evaluate the association between the sensitive features $x_1, \ldots, x_k$ and $x_i$ in the training dataset. It is worth noting that, since the association between some non-sensitive features and the sensitive feature may be trivial, we employ the Wald test with $t$-distribution to check whether the null hypothesis (that the slope $\hat{b}$ of the linear regression model is zero) holds. Specifically, we introduce the $p$-value of the Wald test to avoid unnecessary removing steps, i.e., consider “$p$-value $< 0.05$” as a precondition. If “$p$-value $< 0.05$” holds, we calculate the estimates $\hat{a}_i$ and $\hat{b}_i$ of the Multivariate-regression model, which are sorted in $E_a$ and $E_b$.
        \item The multiple regression model is mathematical as defined in Formula \ref{multiple regression def}:
        % \begin{equation}
        % \label{multiple regression def}
        %    x_i = \beta_0 + \beta_1 \cdot x_1 + \beta_2 \cdot x_2 + \dots + \beta_k \cdot x_k + \epsilon  
        % \end{equation}
        \item The goal is to eliminate these dependencies and generate a new, bias-reduced dataset \(x_{i_{new}}\), such that: with each $i, k+1\leq i\leq d$ then $x_{i_new}$ as defined in Formula \ref{xi_new}
        \begin{equation}
        \label{xi_new}
        x_{i_{new}} = x_i - (\beta_0 + \beta_1 \cdot x_1 + \dots + \beta_k \cdot x_k)    
        \end{equation}
        \item The adjusted training set is then used to train the machine learning model, resulting in a trained model.
    \end{itemize}
    \item Remove Association (Test Set):
    \begin{itemize}
        \item The same multivariate regression is applied to the test set, where associations between sensitive and non-sensitive features are removed before the model is tested. This ensures that the model does not learn biased relationships and can make fair predictions.
    \end{itemize}
    \item Evaluation:
    \begin{itemize}
        \item The trained model is evaluated on the bias-adjusted test set to assess both fairness and performance.
        \item This step is critical to determine whether the removal of bias has maintained or improved the model’s performance and whether it generalizes fairness across various sensitive features.
    \end{itemize}
\end{enumerate}

\subsection{Algorithm}
Based on the multivariate regression model, we propose the FAIREDU fair debugging algorithm, shown in Algorithm \ref{alg: Fairedu}. FAIREDU method includes the following three steps:
\begin{enumerate}
    \item Using multivariate regression, Identify the biased features and estimate their biased parts by evaluating the association between each insensitive feature and all sensitive features (lines 5 to 8)
    \item Exclude the biased parts from the training samples. In this step, for any training sample, we perform the following two operators to remove bias: remove sensitive features (line 16) and modify insensitive feature values (lines 17 to 20)
    \item Apply the same modification on the testing samples (lines 22 to 25), and use SML to predict the label of $x^{te}$ (line 26).
\end{enumerate}
\begin{algorithm}[H]
\caption{Multivariate-regression based FAIREDU} 
\label{alg: Fairedu}
\begin{algorithmic}[1] % Start line numbering at 1
\STATE \textbf{Input:} The training dataset $D_{tr}= \{⟨x_1,y_1⟩, \ldots , ⟨x_n, y_n⟩ \}$, where $x_j = [x^j_1, \ldots , x^j_d]$ is a $d-$dimension vector to denote the $d$ feature values, $ x_1, \ldots, x_k$ are k sensitive features value and the other $x^j_{k+1}, \ldots ,x^j_d$ are non-sensitive feature values, $y_j \in \{0,1\}$ and the testing sample $x^{te}_1, \ldots , x^{te}_d$.

\STATE \textbf{Output:} a ML software $S_{ML}$ and the predicted label $S_{ML}(x^{te})$ for $x^{te}$.

\STATE Initialize $(d-k)-$dimension array $E_a[k+1: d]$ with $E_a[i] = 0$, which is used to store the estimation result of intercept $\hat{a_i}$;

\STATE Initialize k $(d-k)-$dimension arrays $E_{b^1}[k+1: d], \ldots , E_{b^k}[k+1: d]$ with $E_{b^1}[i] = 0, ... , E_{b^k}[i] = 0$, which are used to store the estimation result of intercept \(\hat{b^1_i},..., \hat{b^k_i}\);

\STATE Construct the columns vector $V_1, \ldots , V_k$ of the sensitive feature values from $D_{tr}: V_i=[x^1_i, \ldots, x^n_i]^T, 1\leq i \leq k$;

\FOR{$i\in \{k+1, \ldots, d\}$} 
    \STATE construct the column vector $V_i$ of the current non-sensitive feature values: $V_i=[x^1_i, \ldots , x^n_i]^T$;
    \STATE  apply the linear regression model on $V_i: V_i=a_i+b^1_i\cdot V_1 + \ldots + b^k_i \cdot V_k + \mu$, ; 
    \STATE conduct Wald test with $t-$distribution to get the $p-$value;
    \IF{$p-$value $< 0.05$}
        \STATE estimate $\hat{a_i}$ and $\hat{b^1_i}, \ldots, \hat{b^k_i}$ for $a_i$ and $b^1_i, \ldots, b^k_i$;
        \STATE insert $\hat{a_i}$ and $\hat{b^1_i}, \ldots, \hat{b^k_i}$ into $E_a$ and $E_{b_1}, \ldots, E_{b_k}$: $E_a[i]=\hat{a_i}$, $E_{b_1}[i]=\hat{b_1}, \ldots, E_{b_k}[i]=\hat{b_k}$;
    \ENDIF
\ENDFOR

\FOR{$⟨x_j, y_j⟩ \in D_{tr}$}
    \STATE remove the sensitive feature from $x_j: x_j = x_j [k+1 : d]$;
    \FOR{$i \in {k+1,\ldots,d}$}
        \STATE remove the biased part based on the estimation:\\
        $x^j_i=x^j_i - ( E_a[i]+E_{b_1}[i]\times x^j_1+ \ldots + E_{b_k}[i]\times x^j_k )$
    \ENDFOR
\ENDFOR

\STATE Train ML software $S_{ML}$ from the revised $(d-1)-$dimension training data;

\STATE remove the sensitive feature from $x_{te}: x_{te} = x_{te} [k+1 : d]$;

\FOR{$i \in {k+1, \ldots , d}$}
    \STATE apply the same revision on the testing sample $x_{te}$:\\
    $x^{te}_i=x^{te}_i - ( E_a[i]+E_{b_1}[i]\times x^{te}_1+ \ldots + E_{b_k}[i]\times x^{te}_k )$
\ENDFOR

\RETURN $S_{ML}$ and $S_{ML}(x_{te})$;
\end{algorithmic}
\end{algorithm}

\section{Experiment Setup}
In this section, we describe the data preparation for the experiment as well as the general experiment setup. The details of each dataset, including its name, number of variables/features, total valid data points, list of sensitive variables/ features, and their distribution are shown in Table \ref{Table Dataset}.
\subsection{Selection of Datasets}
\label{subsec: Data}
In this article, we use six popular data sets taken from Kaggle \footnote{Kaggle.com}  and one data set collected from the IT department of Dai Nam University (DNU), Hanoi, Vietnam \footnote{https://dainam.edu.vn/en}. Characters of these seven datasets are presented below.
\begin{enumerate}
    \item \textbf{Adult dataset}. This data set contains 48,842 samples with 14 features. The goal of the data set is to determine whether a person’s annual income can be larger than 50k. This dataset has two sensitive features Gender and race \citep{barry_becker_adult_1996}.
    \item \textbf{COMPAS dataset}. COMPAS is the abbreviation of Correctional Offender Management Profiling for Alternative Sanctions, which is a commercial algorithm for evaluating the possibility of a criminal defendant committing a crime again. The dataset contains the features used by the COMPAS algorithm to score the defendant and the judgment results within two years. There are over 7000 rows in this dataset, with two sensitive features Gender and race \citep{noauthor_propublicacompas-analysis_2024}.
    \item \textbf{Default of Credit Card Clients (Default for short) dataset}. This dataset aims to determine whether customers will default on payment through customers’ information. It contains 30,000 rows and 24 features, including two sensitive features Gender and age \citep{i-cheng_yeh_default_2009}.
    \item \textbf{Predict students' dropout and academic success data set}. This dataset contains data from a higher education institution on various features related to undergraduate students, including demographics, socioeconomic factors, and academic performance, to investigate the impact of these factors on student dropout and academic success. This dataset has two sensitive features Gender and Debtor. It contains 4,425 rows and 34 features.  \citep{noauthor_predict_nodate_dropout}.
    \item \textbf{Student Performance dataset}. This data approaches student achievement in secondary education of two Portuguese schools. The data features include student grades, demographic, social, and school-related features and it was collected by using school reports and questionnaires. This dataset has two sensitive features Gender and Health. It contains 395 rows and 33 features \citep{cortez_student_Performance_2008}.
    \item \textbf{Oulad dataset}. It contains data about courses, students, and their interactions with the Virtual Learning Environment (VLE) for seven selected courses (called modules). Course presentations start in February and October, marked by “B” and “J,” respectively. The dataset consists of tables connected using unique identifiers. All tables are stored in the CSV format. This dataset contains 32,593 rows and 12 features, and it has two sensitive features Gender and Disability \citep{kuzilek_open_2017_Oulad}.
    \item \textbf{DNU dataset}. The data collected spanning over 11 courses, the 11 datasets collected belong to 3 different training programs, so the number of credits for each program and the courses within each program also vary. We have selected similar courses, using equivalent courses to replace different ones. After performing these steps, the new dataset includes 59 features and 411 samples. The normalized dataset consists of 42 features: 6 features about the identity information of students, and 33 features about their score, the remaining 3 features include average score, rating, and prediction labels (safety and risk). All features related to scores are the average scores of courses on a 10-point scale. This dataset has three sensitive features Gender, Birthplace (Zone), and Date of Birth.
\end{enumerate}

\begin{table}[]
%\begin{small}
\begin{center}
\caption{Summary of Datasets}
\label{Table Dataset}
\scalebox{0.8}{ % Giảm kích thước bảng xuống 90%
\begin{tabular}{|p{1cm}|p{2.5cm}|p{1.6cm}|p{1.5cm}|p{1.6cm}|p{2.2cm}|p{1.6cm}|}
\hline

\multicolumn{1}{|c|}{\textbf{N-order}} & \multicolumn{1}{c|}{\textbf{Dataset}}& \multicolumn{1}{c|}{\textbf{\#Feature}} & \multicolumn{1}{c|}{\textbf{Size}} & \multicolumn{1}{c|}{\textbf{Sensitive }} & \multicolumn{2}{c|}{\textbf{Privileged vs.}}\\
& & & & feature & \multicolumn{2}{c|}{\textbf{ Unprivileged}}\\
\hline
&&&  && \cellcolor[HTML]{C0C0C0}Male & \cellcolor[HTML]{C0C0C0}32,650 \\\cline{6-7}
&&&  & \multirow{-2}{*}{Gender} & Female& 16,192\\\cline{5-7}
&&&  && \cellcolor[HTML]{C0C0C0}White& \cellcolor[HTML]{C0C0C0}41761  \\\cline{6-7}
&&&  && Black & 4,685 \\\cline{6-7}
&&&  && Asian-Pac-Islander& 1519  \\\cline{6-7}
&&&  && Amer-Indian-Eskimo& 470\\\cline{6-7}
\multirow{-7}{*}{1}& \multirow{-7}{*}{\textbf{Adult}} & \multirow{-7}{*}{14} & \multirow{-7}{*}{48,842}& \multirow{-5}{*}{Race}& Other & 407\\\hline
&&&  && \cellcolor[HTML]{C0C0C0}Male & \cellcolor[HTML]{C0C0C0}5,819  \\\cline{6-7}
&&&  & \multirow{-2}{*}{Gender} & Female& 1,395 \\\cline{5-7}
&&&  && \cellcolor[HTML]{C0C0C0}Caucasian& \cellcolor[HTML]{C0C0C0}2,454  \\\cline{6-7}
&&&  && African-American  & 3695  \\\cline{6-7}
&&&  && Native American& 20\\\cline{6-7}
&&&  && Asian & 32\\\cline{6-7}
&&&  && Hispanic& 637\\\cline{6-7}
\multirow{-8}{*}{2}& \multirow{-8}{*}{\textbf{Compass}}  & \multirow{-8}{*}{28} & \multirow{-8}{*}{7,214} & \multirow{-6}{*}{Race}& Other & 376\\\hline
&&&  && \cellcolor[HTML]{C0C0C0}Male & \cellcolor[HTML]{C0C0C0}11,888 \\\cline{6-7}
&&&  & \multirow{-2}{*}{Gender} & Female& 18,112\\\cline{5-7}
&&&  && Underage& 17,917\\\cline{6-7}
\multirow{-4}{*}{3}& \multirow{-4}{*}{\textbf{Default}}  & \multirow{-4}{*}{24} & \multirow{-4}{*}{30,000}& \multirow{-2}{*}{Age} & Overage& 12,083\\\hline
&&&  && \cellcolor[HTML]{C0C0C0}Male & \cellcolor[HTML]{C0C0C0}1,557  \\\cline{6-7}
\multirow{-1}{*}{}& \multirow{-1}{*}{\textbf{Student}}&&  & \multirow{-2}{*}{Gender} & Female& 2,868 \\\cline{5-7}
\multirow{-1}{*}{}& \multirow{-1}{*}{\textbf{Dropout}}&&  && \cellcolor[HTML]{C0C0C0}Non-Debtor  & \cellcolor[HTML]{C0C0C0}3,922  \\\cline{6-7}
\multirow{-4}{*}{4}& \multirow{-1}{*}{\textbf{Predict}} & \multirow{-4}{*}{35} & \multirow{-4}{*}{4,425} & \multirow{-2}{*}{Debtor} & Debtor& 503\\\hline
&&&  && \cellcolor[HTML]{C0C0C0}Male & \cellcolor[HTML]{C0C0C0}187 \\\cline{6-7}
&&&  & \multirow{-2}{*}{Gender} & Female& 208\\\cline{5-7}
&\multirow{-2}{*}{\textbf{Student}}&&  && \cellcolor[HTML]{C0C0C0}Verygood(\textless{}=3) & \cellcolor[HTML]{C0C0C0}183 \\\cline{6-7}
\multirow{-4}{*}{5}& \multirow{-2}{*}{\textbf{Performance}}  & \multirow{-4}{*}{33} & \multirow{-4}{*}{395}& \multirow{-2}{*}{Health} & Other (\textgreater{}=4)  & 212\\\hline
&&&  && \cellcolor[HTML]{C0C0C0}Male & \cellcolor[HTML]{C0C0C0}17,875 \\\cline{6-7}
&&&  & \multirow{-2}{*}{Gender} & Female& 14,718\\\cline{5-7}
&&&  && \cellcolor[HTML]{C0C0C0}No& \cellcolor[HTML]{C0C0C0}29,429 \\\cline{6-7}
\multirow{-4}{*}{6}& \multirow{-4}{*}{\textbf{OULAD}} & \multirow{-4}{*}{12} & \multirow{-4}{*}{32,593}& \multirow{-2}{*}{Disability} & Yes& 3,164 \\\hline
&&&  && \cellcolor[HTML]{C0C0C0}Male & \cellcolor[HTML]{C0C0C0}362 \\\cline{6-7}
&&&  & \multirow{-2}{*}{Gender} & Female& 49\\\cline{5-7}
&&&  && \cellcolor[HTML]{C0C0C0}BigCity  & \cellcolor[HTML]{C0C0C0}275 \\\cline{6-7}
&&&  & \multirow{-2}{*}{Zone}& Other & 136\\\cline{5-7}

\multirow{-6}{*}{7}& \multirow{-6}{*}{\textbf{DNU}}& \multirow{-6}{*}{11} & \multirow{-6}{*}{411}& \multirow{-1}{*}{Date of } & \cellcolor[HTML]{C0C0C0}TrueAge  & \cellcolor[HTML]{C0C0C0}281 \\\cline{6-7} 
&&&&Birth&OverAge& 130\\\hline
\end{tabular}    }
\end{center}
%\end{small}
\end{table}
Note: In Table \ref{Table Dataset}, Privilege values are marked with a grey background.

\subsection{Selection of models}
In this paper, we conduct experiments on widely used machine learning models in educational applications, including Logistic Regression, Decision Trees, and Random Forests \citep{kharb_role_2021,luan_review_2021, A_review_2024}.
\begin{itemize}
  
     \item \textbf{\textit{Logistic regression (LR):}} is a statistical method used for binary classification problems, where the goal is to predict one of two possible outcomes. It's a type of regression analysis where the dependent feature is categorical \citep{b5}.
    \item \textbf{\textit{Decision Tree (DT):}} algorithm is a popular machine-learning method for classification and regression tasks. It operates by partitioning the dataset into smaller subsets and constructing a decision tree based on decision rules. Each node in the tree represents a feature, and each edge represents a value of that feature. The leaves of the tree correspond to labels or predicted values \citep{Costa2023}.
    \item \textbf{\textit{Random Forest (RF):}} algorithm is a structured machine-learning approach based on the concept of decision trees. However, instead of using a single decision tree, Random Forest utilizes an ensemble of decision trees, known as a "forest." Each tree in the forest is constructed from a random subset of samples from the training dataset, and features are randomly chosen for each tree during the construction process \citep{Random_Forest}.
\end{itemize}

\subsection{Evaluation Metrics}
\label{subsubsec: Evaluation Metrics}
A number of fairness metrics are widely used in AI fairness research \citep{b2,b3,b8,b15,b32}, including Disparate Impact (DI), Statistical Parity Difference (SPD), Average Odds Difference (AOD), and Equal Opportunity Difference (EOD). Descriptions of these metrics are given in Table \ref{Table Fairness Metrics}. 

\begin{table}[]
\centering
\caption{Type of Fairness Metrics}
\label{Table Fairness Metrics}
    \begin{tabular}{|p{3cm}|p{5cm}|p{5cm}|}
    \hline
       \textbf{Fairness Metrics} & \textbf{Description} &\textbf{Math} \\
    \hline 
    \textbf{Disparate Impact (DI)} & The ratio of the favorable rate of the unprivileged group to the favorable rate of the privileged group & \( DI=\dfrac{p(\hat{y}=1|A=0)}{p(\hat{y}=1|A=1)} \)\\
\hline
\textbf{Statistical Parity Difference (SPD)} & The disparity in favorable rates between the privileged and unprivileged groups& \( SPD=p(\hat{y}=1|A=0) - p(\hat{y}=1|A=1) \)\\
\hline
\textbf{AOD (Average Odds Difference)} & The average discrepancy between privileged and unprivileged groups between false-positive and true-positive rates& \( AOD=\dfrac{1}{2}(p(\hat{y}=1|A=0,y=0) - p(\hat{y}=1|A=1,y=0)+p(\hat{y}=1|A=0,y=1) - p(\hat{y}=1|A=1,y=1)) \)\\
\hline
\textbf{EOD (Equal Opportunity Difference)} & The disparity in true-positive rates between the privileged and unprivileged groups & \( EOD=p(\hat{y}=1|A=0,y=1) - p(\hat{y}=1|A=1,y=1)) \)\\
\hline
    \end{tabular}
   
\end{table}

We adopt all of these metrics to capture a comprehensive view of fairness, as each metric focuses on different aspects of bias in machine learning outcomes. By using a variety of fairness metrics, such as DI, SPD, AOD, and EOD, we ensure that the evaluation considers both group-level disparities and individual-level prediction fairness.

To investigate the trade-off between fairness and performance, we also evaluate the performance of models using the two most popular metrics, which are Accuracy (ACC) in Formula \ref{ACC} and Recall in Formula \ref{Recall} \cite{acc_and_recall}.
\begin{equation}
\label{ACC}
    ACC=\dfrac{TP+TN}{Total}
\end{equation}

\begin{equation}
\label{Recall}
    Recall = \dfrac{TP}{TP+FN}
\end{equation}
   
Where $TP$ denotes the true positive samples, $TN$ denotes the true negative samples, and $Total$ denotes the total samples. $FN$ denotes the false negative samples.

\section{Results}
\label{sec: Results}
This section presents the experimental results aimed at addressing all the research questions outlined in \label{sec: Intro}, including RQ1 (Section 5.1), RQ2 (Section 5.2), RQ3 (Section 5.3), and RQ4 (Section 5.4).

\newtcolorbox[auto counter, number within=section]{note}[2][]{%
    colback=yellow!10, colframe=yellow!80!black, 
    fonttitle=\bfseries, 
    title=Note~\thetcbcounter: #2, #1
}
\subsection{RQ1 -  Is there a systematic bias present among sensitive features within educational datasets?}

\begin{figure}[htbp]
\centerline{\includegraphics[width=0.8\textwidth]{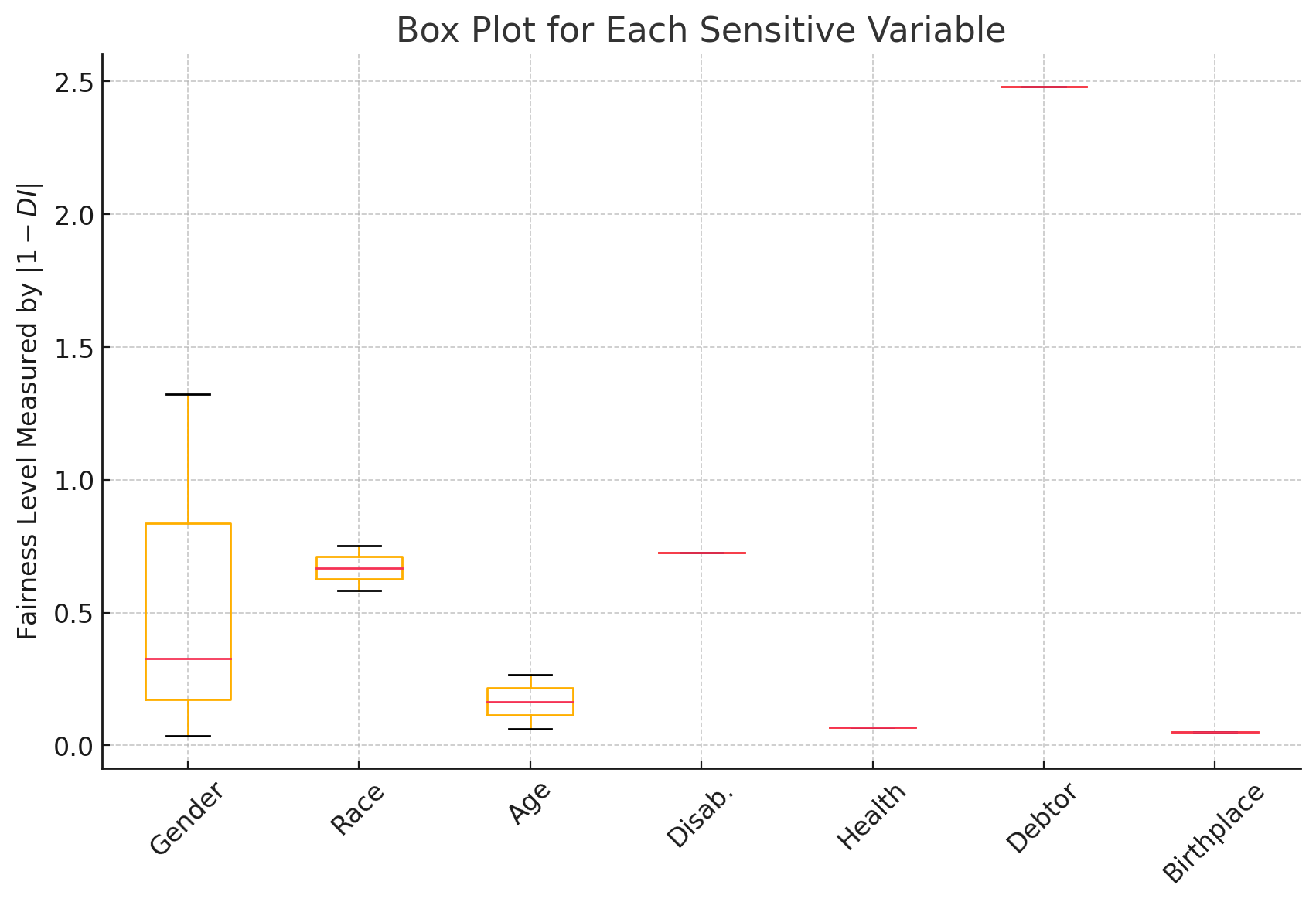}}
\caption{Comparing $|1-DI|$ of sensitive features across datasets}
\label{Fig. 2}
\end{figure}

The results of fairness levels, measured by $|1-DI|$, for various sensitive features, including Gender, Race, Age, Disability (Disab.), Health, Debtor, and Birthplace were presented in Figure \ref{Fig. 2} . The figure shows that the Gender feature shows the widest range of values, indicating significant variability in fairness across different contexts or datasets. The fairness value of Race and Age does not vary much. We can only collect one value point for each feature: Disability, Health, Debtor, and Birthplace. However, it can be seen that there are no patterns regarding the order of biasness among these sensitive features. To ensure a thorough evaluation of fairness, it is essential to take into account all sensitive features present in the dataset.
\label{subsec: RQ1}
\begin{note}[label={note:1}]{Sensitive features in the educational datasets}
The analysis reveals no consistent bias across sensitive features within the educational datasets. In other words, no single sensitive feature consistently demonstrates greater unfairness than the others.
   
\end{note}

\subsection{RQ2 - Does the level of fairness vary across different machine learning methods?}
\label{subsec: RQ2}

Table \ref{tab:Table 5} displayed the average fairness value across seven datasets for each sensitive feature with each ML method. We present four figures according to four fairness metrics, which are $|1-DI|$, SPD, AOD, and EOD.
In the first figure, we compared the $|1-DI|$ value among three ML models: logistic regression, random forest, and decision tree. The result shows that overall logistic regression leads to a higher level of bias across the dataset. This observation is the same with different measures of fairness, including SPD, AOD, and EOD in the next figures. Decision Tree is the model with the lowest level of bias across different sensitive features.  

Logistic regression tends to be more sensitive to the presence of biased data because it applies the same linear weights across all instances. If the training data reflects historical biases or unequal distributions, the model will inherently reproduce and potentially amplify these biases. Moreover, the linear nature of logistic regression makes it prone to capturing and amplifying relationships between sensitive features, such as gender or race, and the target feature, leading to unfair outcomes among different groups. For example, if "gender" strongly influences the outcome, this model will reflect that difference.

\begin{table}[H]
    \centering
     \caption{Comparing Fairness measures for sensitive features in different ML algorithms}
    \label{tab:Table 5} 
    \begin{tabular}{|c|c|} \hline 
    \includegraphics[width=0.5\linewidth]{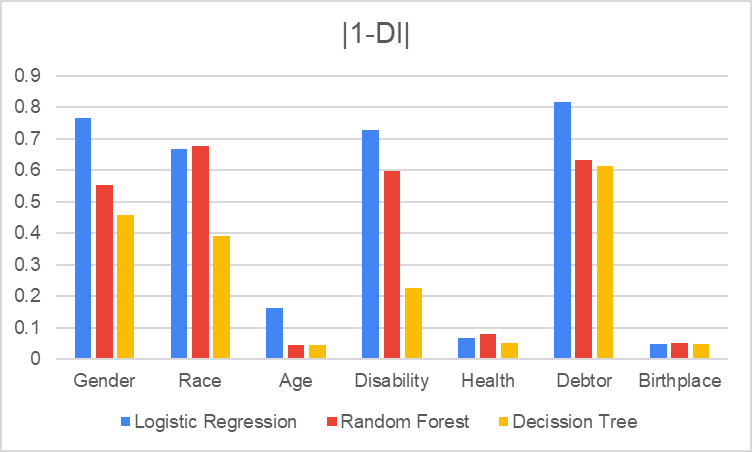}
 &     \includegraphics[width=0.5\linewidth]{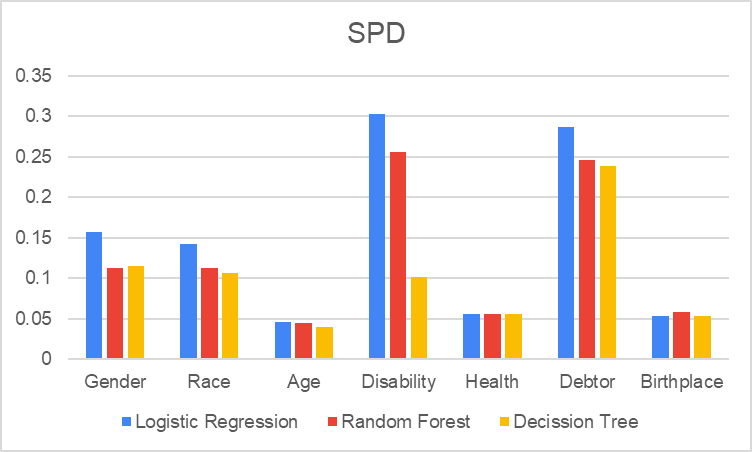}\\ \hline 
          \includegraphics[width=0.5\linewidth]{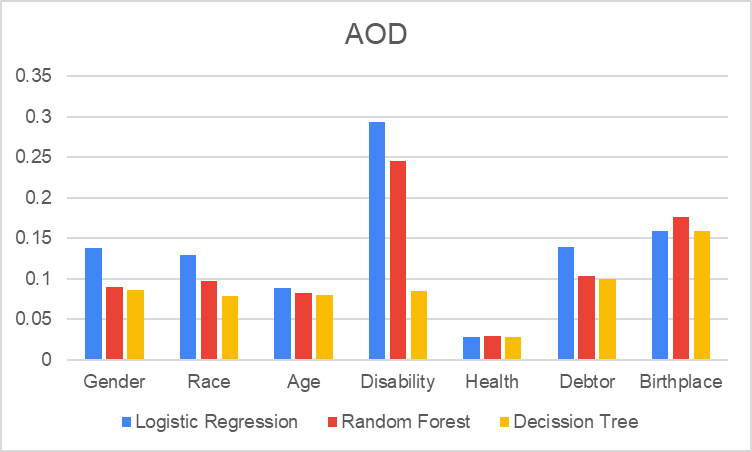}   &     \includegraphics[width=0.5\linewidth]{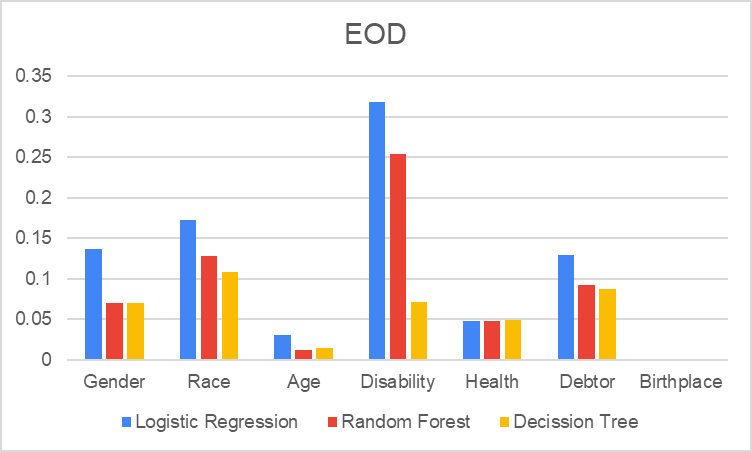}\\ \hline 
              \end{tabular}
  
\end{table}

Decision trees split data at each node based on the optimal feature threshold that maximizes information gain (or minimizes impurity), considering local patterns. This flexibility allows decision trees to capture non-linear relationships and adapt to different contexts within the data. As a result, decision trees can better handle complex scenarios where biases might manifest differently in various subsets of the data, leading to lower overall bias levels. 

Random forests, by using multiple decision trees, can offer similar or even better fairness, as they mitigate the influence of bias if any individual tree is skewed by a sensitive feature.

We also compare fairness measures. with $|1-DI|$, debtor is the feature with the highest level of bias. However, with SPD, AOD, and EOD, disability is the feature with the highest level of bias. 
The debtor status shows the highest bias under $|1-DI|$ likely because The "debtor" feature shows the highest level of bias because this measure is sensitive to the disparity in the proportion of positive outcomes between groups. In this case, the proportion of debtors (503/4425) is significantly lower compared to non-debtors, and the large difference in positive outcomes between these groups leads to a higher $|1-DI|$. One note that, "debtor" only appears in the Student-Dropout-Predict dataset.
% This measure is sensitive to the distribution of outcomes between groups. If debtors are systematically less likely to receive a favorable outcome compared to non-debtors, 
%will highlight this stark imbalance. The nature of financial data, where historical lending or credit decisions may have been more stringent against debtors, could lead to a pronounced disparate impact for this feature.

On the other hand, with SPD, AOD, and EOD, the "disability" feature exhibits the highest bias. This is likely due to the significant disparity in the ratio of disabled and non-disabled individuals (3164/32594), and the considerable difference in the model’s ability to correctly predict positive outcomes for these two groups.\\
This highlights that different fairness metrics capture different aspects of fairness. $|1-DI|$ focuses on the distribution of positive outcomes across groups, while SPD evaluates the difference in positive prediction rates between groups, without considering accuracy. AOD and EOD, however, consider the true positive rate (TPR) and true negative rate (TNR), reflecting how balanced the model’s predictions are across different groups. The disability feature shows the highest bias under SPD, AOD, and EOD because these measures capture different types of biases that go beyond simple outcome distributions:
\begin{itemize}
    \item SPD (Statistical Parity Difference): Indicates a disparity in the overall likelihood of receiving a positive prediction between groups. If individuals with disabilities are less likely to receive positive outcomes regardless of their actual qualifications, SPD will detect this bias.
    \item AOD (Average Odds Difference): Evaluates the difference in error rates (false positives and false negatives) between groups. If a model is more likely to misclassify individuals with disabilities, this would lead to a high AOD.
    \item EOD (Equal Opportunity Difference): Focuses on the difference in true positive rates between groups. If individuals with disabilities who qualify for a positive outcome (e.g., job suitability or creditworthiness) are less likely to actually receive it, EOD will be high.

\end{itemize}

Alternatively, Table \ref{tab:Table 5} demonstrates that, for the same model, different fairness metrics yield varying results. For instance, the Disability feature exhibits greater fairness than the Debtor feature when evaluated using Disparate Impact (DI) and Statistical Parity Difference (SPD). However, when assessed through Average Odds Difference (AOD) and Equal Opportunity Difference (EOD), the Disability feature is found to be less fair.

Disparate Impact (DI) and Statistical Parity Difference (SPD) assess fairness by comparing the percentage of favorable outcomes between groups, without considering the accuracy of predictions. In contrast, Average Odds Difference (AOD) and Equal Opportunity Difference (EOD) focus on the quality of predictions, evaluating fairness based on true positive or true negative rates across groups. As a result, optimizing fairness according to one metric can potentially compromise fairness according to another. For instance, improving fairness in terms of Equal Opportunity may require lowering the model’s overall accuracy by adjusting decision thresholds to equalize true positive rates between groups.

This highlights the importance of selecting fairness measures that align with the specific context and goals of the analysis. If the objective is to ensure an equitable distribution of favorable outcomes, metrics like $|1-DI|$ or SPD would be appropriate. However, if the emphasis is on the accuracy of positive predictions for different groups, AOD and EOD offer a more meaningful evaluation. Given the potential trade-offs between these measures, the choice of fairness metric should be guided by the particular fairness goals in a given scenario.

% Define a custom tcolorbox style

% Use the custom note box
\begin{note}[label={note:2}]{Fairness different across different ML methods}
    \begin{itemize}
        \item There are differences in bias level for different ML models. The LR model shows a greater risk of bias than the RF and DT models.
        \item The Order of fairness level for different sensitive features differs for different fairness measures. If the focus is on the distribution of favorable outcomes across groups, DI and SPD serve as appropriate metrics. Contrary, if the objective is to examine the balance of predictions across outcome groups, AOD and EOD are more suitable.
        \item To draw comprehensive conclusions about fairness, it is crucial to consider multiple metrics.
    \end{itemize}
\end{note}
\begin{comment}
We estimate the average of fairness measures for three models, including Logistic Regression, Random Forest, and Decision Tree. The results are illustrated in Figure \ref{Fig. Aver_of_Fainess}\\
\begin{figure}[htbp]
\centerline{\includegraphics[width=0.9\textwidth]{Figure/Aver_of_Fairness.png}}
\caption{\centering{Averange of the Fairness Measure}\\
\textit{Note: All fairness measure ($|1-DI|$, SPD, AOD, EOD) as small as good}}
\label{Fig. Aver of Fainess}
\end{figure}
    The results of measuring the fairness of the sensitive features across the models show that the age-sensitive features are quite stable across the models and the fairness measure $|1-DI|$ is quite small, fluctuating below 0.2. However, the features of Gender and Race show a large degree of fluctuation across the datasets and the fairness measure $|1-DI|$ fluctuates around the threshold of 0.7, indicating a rather low fairness.

\begin{figure}[htbp]

\centerline{\includegraphics[width=0.9\textwidth]{Figure/Av_of_1-DI.png}}
\caption{Average of $|1-DI|$ of Models}
\label{Fig. Av |1-DI|}
\end{figure}
\end{comment}

\subsection{RQ3 - How does FAIREDU manage multiple sensitive features compared to current state-of-the-art methods?}
\label{subsec: RQ3}
To assess the fairness improvement of the FAIREDU method, we compared it against other state-of-the-art methods such as Reweighing, DIR, Fairway, FairSmote, and LTDD \citep{b27} across multiple machine learning models, including Logistic Regression, Random Forest, and Decision Tree. Table \ref{tab: Compare fairness Origin} presents the comparison across different methods and datasets presented in LTDD study \citep{b27}. The comparison results with the original model and other state-of-the-art models show that FAIREDU outperforms in most cases across the Adult, COMPAS, Default, and Student datasets. However, for $|1-DI|$ on the Compas$\_$sex and Default$\_$sex features, we fall slightly behind LTDD, but the difference is not significant (less than 0.1). Similarly, for SPD, our results are only marginally lower than LTDD, with a difference of less than 0.01.
%Bổ sung bảng so sánh với các model khác - trích từ bản gốc LTDD

% So sánh mức độ tăng của fairness cho từng dataset khi áp dụng edufair

% So sánh mức độ tăng của fairness cho các fairness metrics khác nhau khi áp dụng edufair

% So sánh mức độ tăng của fairness đối với các model khác nhau khi áp dụng edufair

% So sánh performance giữa LTDD và EduFair đối với từng biến

% Please add the following required packages to your document preamble:
% \usepackage{multirow}
% \usepackage[table,xcdraw]{xcolor}
% Beamer presentation requires \usepackage{colortbl} instead of \usepackage[table,xcdraw]{xcolor}

\begin{table}[H]

\begin{small}
\begin{center}
\caption{Comparing FAIREDU with existing methodes (Table extended from \citep{b27}). Gray boxes indicate results that outperform the baseline (FAIREDU). Black boxes indicate results that are worse than the baseline (FAIREDU).
}
\label{tab: Compare fairness Origin}
\scalebox{0.8}{ % Giảm kích thước bảng xuống 80%
\begin{tabular}{|p{2cm}|p{2cm}|p{1cm}|p{1cm}|p{1.4cm}|p{1.4cm}|p{1cm}|p{1cm}|p{1.5cm}|} 
\hline 
\multicolumn{1}{|l|}{\textbf{Indicators}} & \textbf{Method} & \multicolumn{1}{p{1cm}|}{\textbf{Adult Race}} & \multicolumn{1}{p{1cm}|}{\textbf{Adult Sex}} & \multicolumn{1}{p{1.4cm}|}{\textbf{Compas Race}} & \multicolumn{1}{p{1.4cm}|}{\textbf{Compas Sex}} & \multicolumn{1}{c|}{\textbf{Default}} & \multicolumn{1}{c|}{\textbf{Student}}&\textbf{W/T/L} \\ 
\hline

& Original   & \cellcolor[HTML]{C0C0C0}0.5894 & \cellcolor[HTML]{C0C0C0}0.8531 & \cellcolor[HTML]{C0C0C0}0.7929 & \cellcolor[HTML]{C0C0C0}1.3061 & \cellcolor[HTML]{C0C0C0}0.3139 & \cellcolor[HTML]{C0C0C0}0.1705 &\textbf{6/0/0}\\ 
\cdashline{2-9} 

& Reweighing & \cellcolor[HTML]{C0C0C0}0.2744 & \cellcolor[HTML]{C0C0C0}0.4533 & \cellcolor[HTML]{C0C0C0}0.1244 & \cellcolor[HTML]{C0C0C0}0.1278 & \cellcolor[HTML]{343434}{\color[HTML]{FFFFFF} 0.0866} & \cellcolor[HTML]{C0C0C0}0.1387 &\textbf{5/0/1}\\ 
\cdashline{2-9} 

& DIR        & \cellcolor[HTML]{C0C0C0}0.6837 & \cellcolor[HTML]{C0C0C0}0.8787 & \cellcolor[HTML]{C0C0C0}0.8261 & \cellcolor[HTML]{C0C0C0}1.3117 & \cellcolor[HTML]{C0C0C0}0.274  & \cellcolor[HTML]{C0C0C0}0.1635 &\textbf{6/0/0}\\ 
\cdashline{2-9} 

& Fairway    & \cellcolor[HTML]{C0C0C0}0.5099 & \cellcolor[HTML]{C0C0C0}nan*   & \cellcolor[HTML]{C0C0C0}0.5639 & \cellcolor[HTML]{C0C0C0}1.6904 & \cellcolor[HTML]{C0C0C0}0.3071 & \cellcolor[HTML]{C0C0C0}0.1903 &\textbf{6/0/0}\\ 
\cdashline{2-9} 

& Fair-Smote & \cellcolor[HTML]{C0C0C0}0.2184 & \cellcolor[HTML]{C0C0C0}0.2655 & \cellcolor[HTML]{C0C0C0}0.0801 & \cellcolor[HTML]{C0C0C0}0.0851 & \cellcolor[HTML]{343434}{\color[HTML]{FFFFFF} 0.0665}& \cellcolor[HTML]{C0C0C0}0.1811 &\textbf{5/0/1}\\ 
\cdashline{2-9} 

& LTDD       & \cellcolor[HTML]{C0C0C0}0.2027 & \cellcolor[HTML]{C0C0C0}0.2136 & \cellcolor[HTML]{C0C0C0}0.1381 & \cellcolor[HTML]{343434}{\color[HTML]{FFFFFF} 0.079} & \cellcolor[HTML]{343434}{\color[HTML]{FFFFFF} 0.085} & \cellcolor[HTML]{C0C0C0}0.1686 &\textbf{4/0/2}\\ 
\cline{2-9} 
\multirow{-7}{*}{\textbf{$|1-DI|$}} 
& FairEdu    & \multicolumn{1}{r|}{\cellcolor[HTML]{FFFFFF}{\color[HTML]{212121} 0.172}} & \multicolumn{1}{r|}{\cellcolor[HTML]{FFFFFF}{\color[HTML]{212121} 0.162}} & \multicolumn{1}{r|}{\cellcolor[HTML]{FFFFFF}{\color[HTML]{212121} 0.030}} & \multicolumn{1}{r|}{\cellcolor[HTML]{FFFFFF}{\color[HTML]{212121} 0.084}} & \multicolumn{1}{r|}{\cellcolor[HTML]{FFFFFF}{\color[HTML]{212121} 0.177}} & \multicolumn{1}{r|}{\cellcolor[HTML]{FFFFFF}{\color[HTML]{212121} 0.128}} &\\ 
\hline
\hline

& Original   & \cellcolor[HTML]{C0C0C0}0.0899 & \cellcolor[HTML]{C0C0C0}0.1659 & \cellcolor[HTML]{C0C0C0}0.2037 & \cellcolor[HTML]{C0C0C0}0.2605 & \cellcolor[HTML]{C0C0C0}0.0279 & \cellcolor[HTML]{343434}{\color[HTML]{FFFFFF} 0.0714}&\textbf{5/0/1} \\ 
\cdashline{2-9} 

& Reweighing & \cellcolor[HTML]{C0C0C0}0.04   & \cellcolor[HTML]{C0C0C0}0.0653 & \cellcolor[HTML]{C0C0C0}0.0535 & \cellcolor[HTML]{C0C0C0}0.0494 & \cellcolor[HTML]{343434}{\color[HTML]{FFFFFF} 0.0064} & \cellcolor[HTML]{343434}{\color[HTML]{FFFFFF} 0.0583}&\textbf{4/0/2} \\ 
\cdashline{2-9} 

& DIR        & \cellcolor[HTML]{C0C0C0}0.1383 & \cellcolor[HTML]{C0C0C0}0.2101 & \cellcolor[HTML]{C0C0C0}0.2061 & \cellcolor[HTML]{C0C0C0}0.2604 & \cellcolor[HTML]{C0C0C0}0.0226 & \cellcolor[HTML]{343434}{\color[HTML]{FFFFFF} 0.0679} &\textbf{5/0/1}\\ 
\cdashline{2-9} 

& Fairway    & \cellcolor[HTML]{C0C0C0}0.0598 & \cellcolor[HTML]{343434}{\color[HTML]{FFFFFF} 0.0018} & \cellcolor[HTML]{C0C0C0}0.1828 & \cellcolor[HTML]{C0C0C0}0.2956 & \cellcolor[HTML]{C0C0C0}0.0254 & \cellcolor[HTML]{C0C0C0}0.0793 &\textbf{5/0/1}\\ 
\cdashline{2-9} 

& Fair-Smote & \cellcolor[HTML]{C0C0C0}0.0789 & \cellcolor[HTML]{C0C0C0}0.1005 & \cellcolor[HTML]{C0C0C0}0.0372 & \cellcolor[HTML]{343434}{\color[HTML]{FFFFFF} 0.0399} & \cellcolor[HTML]{C0C0C0}0.0211 & \cellcolor[HTML]{343434}{\color[HTML]{FFFFFF} 0.0741}&\textbf{4/0/2} \\ 
\cdashline{2-9} 

& LTDD       & \cellcolor[HTML]{C0C0C0}0.0293 & \cellcolor[HTML]{C0C0C0}0.0272 & \cellcolor[HTML]{C0C0C0}0.0616 & \cellcolor[HTML]{343434}{\color[HTML]{FFFFFF} 0.0347} & \cellcolor[HTML]{343434}{\color[HTML]{FFFFFF} 0.0059} & \cellcolor[HTML]{343434}{\color[HTML]{FFFFFF} 0.07} &\textbf{3/0/3}\\ 
\cline{2-9} 

\multirow{-7}{*}{\textbf{SPD}} 
& FairEdu    & \multicolumn{1}{r|}{\cellcolor[HTML]{FFFFFF}{\color[HTML]{212121} 0.020}} & \multicolumn{1}{r|}{\cellcolor[HTML]{FFFFFF}{\color[HTML]{212121} 0.019}} & \multicolumn{1}{r|}{\cellcolor[HTML]{FFFFFF}{\color[HTML]{212121} 0.028}} & \multicolumn{1}{r|}{{\color[HTML]{212121} 0.046}} & \multicolumn{1}{r|}{{\color[HTML]{212121} 0.014}} & \multicolumn{1}{r|}{{\color[HTML]{212121} 0.076}}& \\ 
\hline

\end{tabular}}
\end{center}
\end{small}
\end{table}

%The experiment above is repeated for 
%We conducted comprehensive experiments on the Adult, COMPAS, Default, Oulad, Student Performance, Student Dropout, and DNU datasets, comparing the performance of both the LTDD and FAIREDU methods. The detailed result of Logistic Regression is presented in Tables \ref{tab: Compare Bef-Af Logistic Regression}, and the results of the RF and DT model were summarised in Figure \ref{SPD_RF} and \ref{1-DI_DT}.\\

%\begin{landscape}

\begin{table}[]

\begin{center}
\caption{Pretest-postest comparison for LTDD and FAIREDU with different fairness measures using Logistic Regression models. Gray boxes indicate results that outperform FAIREDU. Black boxes indicate results that are worse than FAIREDU. Percent change compared against the Original dataset}
\label{tab: Compare Bef-Af Logistic Regression}
\begin{small}
\scalebox{0.6}{ % Giảm kích thước bảng xuống 90%
    \begin{tabular}{|p{1.5cm}||p{1.2cm}|p{1.7cm}|p{1.7cm}|p{1.2cm}|p{1.7cm}|p{1.7cm}|p{1.2cm}|p{1.7cm}|p{1.7cm}|p{1.2cm}|p{1.7cm}|p{1.7cm}|}
    \hline
         \textbf{Dataset/}&   \multicolumn{3}{|c|}{$|1-DI|$}&  \multicolumn{3}{|c|}{SPD  }&\multicolumn{3}{|c|}{AOD}& \multicolumn{3}{|c|}{EOD}\\ 
         \cline{2-13} 
         \textbf{S.Var.}& Origin.&LTDD (\%change)&  FairEdu (\%change)&  Orign.&LTDD (\%change)&  FairEdu (\%change)& Orign.&LTDD (\% change)&  FairEdu (\%change)& Orign.&LTDD (\%change)&  FairEdu (\%change)\\ \hline 
     \multirow{-1}{*}{Adult}   & \cellcolor[HTML]{C0C0C0}{\color[HTML]{000000} }    & \cellcolor[HTML]{C0C0C0}{\color[HTML]{000000} 0.219}      & 0.162      & \cellcolor[HTML]{C0C0C0}     & \cellcolor[HTML]{C0C0C0}0.028    & 0.019      & \cellcolor[HTML]{C0C0C0}     & \cellcolor[HTML]{343434}{\color[HTML]{FFFFFF} 0.050}      & 0.059      & \cellcolor[HTML]{C0C0C0}    & \cellcolor[HTML]{343434}{\color[HTML]{FFFFFF} 0.078}      & 0.092      \\
  \multirow{-1}{*}{Gender}     & \multirow{-2}{*}{\cellcolor[HTML]{C0C0C0}{\color[HTML]{000000} 0.856}} & \cellcolor[HTML]{C0C0C0}{\color[HTML]{000000} (-74.42\%)} & (-81.07\%) & \multirow{-2}{*}{\cellcolor[HTML]{C0C0C0}0.166}    & \cellcolor[HTML]{C0C0C0}(-83.13\%)    & (-88.55\%) & \multirow{-2}{*}{\cellcolor[HTML]{C0C0C0}0.172}    & \cellcolor[HTML]{343434}{\color[HTML]{FFFFFF} (-70.93\%)} & (-65.7\%)  & \multirow{-2}{*}{\cellcolor[HTML]{C0C0C0}0.273} & \cellcolor[HTML]{343434}{\color[HTML]{FFFFFF} (-71.43\%)} & (-66.3\%)  \\\hline
 \multirow{-1}{*}{Adult}     & \cellcolor[HTML]{C0C0C0}{\color[HTML]{000000} }    & \cellcolor[HTML]{C0C0C0}{\color[HTML]{000000} 0.201}      & 0.172      & \cellcolor[HTML]{C0C0C0}     & \cellcolor[HTML]{C0C0C0}0.029    & 0.020      & \cellcolor[HTML]{C0C0C0}     & \cellcolor[HTML]{C0C0C0}0.030    & 0.025      & \cellcolor[HTML]{C0C0C0}    & \cellcolor[HTML]{C0C0C0}0.062    & 0.049      \\
 \multirow{-1}{*}{Race}       & \multirow{-2}{*}{\cellcolor[HTML]{C0C0C0}{\color[HTML]{000000} 0.583}} & \cellcolor[HTML]{C0C0C0}{\color[HTML]{000000} (-65.52\%)} & (-70.5\%)  & \multirow{-2}{*}{\cellcolor[HTML]{C0C0C0}0.088}    & \cellcolor[HTML]{C0C0C0}(-67.05\%)    & (-77.27\%) & \multirow{-2}{*}{\cellcolor[HTML]{C0C0C0}0.079}    & \cellcolor[HTML]{C0C0C0}(-62.03\%)    & (-68.35\%) & \multirow{-2}{*}{\cellcolor[HTML]{C0C0C0}0.118} & \cellcolor[HTML]{C0C0C0}(-47.46\%)    & (-58.47\%) \\\hline
\multirow{-1}{*}{Compas}    & \cellcolor[HTML]{C0C0C0}{\color[HTML]{000000} }    & \cellcolor[HTML]{343434}{\color[HTML]{FFFFFF} 0.023}     & 0.084      & \cellcolor[HTML]{C0C0C0}     & \cellcolor[HTML]{343434}{\color[HTML]{FFFFFF} 0.037}      & 0.046      & \cellcolor[HTML]{C0C0C0}     & \cellcolor[HTML]{343434}{\color[HTML]{FFFFFF} 0.046}      & 0.076      & \cellcolor[HTML]{C0C0C0}    & \cellcolor[HTML]{343434}{\color[HTML]{FFFFFF} 0.054}      & 0.086      \\
 \multirow{-1}{*}{Gender}     & \multirow{-2}{*}{\cellcolor[HTML]{C0C0C0}{\color[HTML]{000000} 1.323}} &\cellcolor[HTML]{343434}{\color[HTML]{FFFFFF} (-98.26\%)}   & (-93.65\%) & \multirow{-2}{*}{\cellcolor[HTML]{C0C0C0}0.263}    & \cellcolor[HTML]{343434}{\color[HTML]{FFFFFF} (-85.93\%)} & (-82.51\%) & \multirow{-2}{*}{\cellcolor[HTML]{C0C0C0}0.243}    & \cellcolor[HTML]{343434}{\color[HTML]{FFFFFF} (-81.07\%)} & (-68.72\%) & \multirow{-2}{*}{\cellcolor[HTML]{C0C0C0}0.282} & \cellcolor[HTML]{343434}{\color[HTML]{FFFFFF} (-80.85\%)} & (-69.5\%)  \\\hline
\multirow{-1}{*}{Compas}     & \cellcolor[HTML]{C0C0C0}     & \cellcolor[HTML]{C0C0C0}0.066    & 0.030      & \cellcolor[HTML]{C0C0C0}     & \cellcolor[HTML]{C0C0C0}0.063    & 0.028      & \cellcolor[HTML]{C0C0C0}     & \cellcolor[HTML]{C0C0C0}0.069    & 0.027      & \cellcolor[HTML]{C0C0C0}    & \cellcolor[HTML]{C0C0C0}0.063    & 0.049      \\
  \multirow{-1}{*}{Race}       & \multirow{-2}{*}{\cellcolor[HTML]{C0C0C0}0.752}    & \cellcolor[HTML]{C0C0C0}(-91.22\%)    & (-96.01\%) & \multirow{-2}{*}{\cellcolor[HTML]{C0C0C0}0.197}    & \cellcolor[HTML]{C0C0C0}(-68.02\%)    & (-85.79\%) & \multirow{-2}{*}{\cellcolor[HTML]{C0C0C0}0.181}    & \cellcolor[HTML]{C0C0C0}(-61.88\%)    & (-85.08\%) & \multirow{-2}{*}{\cellcolor[HTML]{C0C0C0}0.228} & \cellcolor[HTML]{C0C0C0}(-72.37\%)    & (-78.51\%) \\\hline
\multirow{-1}{*}{Default}    & \cellcolor[HTML]{C0C0C0}     & \cellcolor[HTML]{C0C0C0}0.159    & 0.028      & \cellcolor[HTML]{C0C0C0}     & \cellcolor[HTML]{C0C0C0}0.014    & 0.008      & \cellcolor[HTML]{C0C0C0}     & \cellcolor[HTML]{C0C0C0}0.019    & 0.013      & \cellcolor[HTML]{C0C0C0}    & \cellcolor[HTML]{C0C0C0}0.036    & 0.024      \\
 \multirow{-1}{*}{Age}        & \multirow{-2}{*}{\cellcolor[HTML]{C0C0C0}0.266}    & \cellcolor[HTML]{C0C0C0}(-40.23\%)    & (-89.47\%) & \multirow{-2}{*}{\cellcolor[HTML]{C0C0C0}0.023}    & \cellcolor[HTML]{C0C0C0}(-39.13\%)    & (-65.22\%) & \multirow{-2}{*}{\cellcolor[HTML]{C0C0C0}0.034}    & \cellcolor[HTML]{C0C0C0}(-44.12\%)    & (-61.76\%) & \multirow{-2}{*}{\cellcolor[HTML]{C0C0C0}0.061} & \cellcolor[HTML]{C0C0C0}(-40.98\%)    & (-60.66\%) \\\hline
 \multirow{-1}{*}{Default}    & \cellcolor[HTML]{C0C0C0}     & \cellcolor[HTML]{343434}{\color[HTML]{FFFFFF} 0.025}     & 0.177      & \cellcolor[HTML]{C0C0C0}     & \cellcolor[HTML]{343434}{\color[HTML]{FFFFFF} 0.006}      & 0.014      & \cellcolor[HTML]{C0C0C0}     & \cellcolor[HTML]{C0C0C0}0.015    & 0.010      & \cellcolor[HTML]{C0C0C0}    & \cellcolor[HTML]{C0C0C0}0.032    & 0.017      \\
  \multirow{-1}{*}{Gender}     & \multirow{-2}{*}{\cellcolor[HTML]{C0C0C0}0.327}    & \cellcolor[HTML]{343434}{\color[HTML]{FFFFFF} (-92.35\%)}   & (-45.87\%) & \multirow{-2}{*}{\cellcolor[HTML]{C0C0C0}0.029}    & \cellcolor[HTML]{343434}{\color[HTML]{FFFFFF} (-79.31\%)} & (-51.72\%) & \multirow{-2}{*}{\cellcolor[HTML]{C0C0C0}0.031}    & \cellcolor[HTML]{C0C0C0}(-51.61\%)    & (-67.74\%) & \multirow{-2}{*}{\cellcolor[HTML]{C0C0C0}0.048} & \cellcolor[HTML]{C0C0C0}(-33.33\%)    & (-64.58\%) \\\hline
 \multirow{-1}{*}{Oulad}     & \cellcolor[HTML]{C0C0C0}     & \cellcolor[HTML]{C0C0C0}0.039    & 0.010      & \cellcolor[HTML]{C0C0C0}     & \cellcolor[HTML]{C0C0C0}0.018    & 0.012      & \cellcolor[HTML]{C0C0C0}     & \cellcolor[HTML]{C0C0C0}0.021    & 0.013      & \cellcolor[HTML]{C0C0C0}    & \cellcolor[HTML]{C0C0C0}0.019    & 0.018      \\
    \multirow{-1}{*}{Gender}     & \multirow{-2}{*}{\cellcolor[HTML]{C0C0C0}0.306}    & \cellcolor[HTML]{C0C0C0}(-87.25\%)    & (-96.73\%) & \multirow{-2}{*}{\cellcolor[HTML]{C0C0C0}0.104}    & \cellcolor[HTML]{C0C0C0}(-82.69\%)    & (-88.46\%) & \multirow{-2}{*}{\cellcolor[HTML]{C0C0C0}0.100}    & \cellcolor[HTML]{C0C0C0}(-79.\%) & (-87.\%)   & \multirow{-2}{*}{\cellcolor[HTML]{C0C0C0}0.119} & \cellcolor[HTML]{C0C0C0}(-84.03\%)    & (-84.87\%) \\\hline
 \multirow{-1}{*}{Oulad}     & \cellcolor[HTML]{C0C0C0}     & \cellcolor[HTML]{343434}{\color[HTML]{FFFFFF} 0.015}     & 0.025      & \cellcolor[HTML]{C0C0C0}     & 0.022   & 0.022      & \cellcolor[HTML]{C0C0C0}     & \cellcolor[HTML]{343434}{\color[HTML]{FFFFFF} 0.029}      & 0.031      & \cellcolor[HTML]{C0C0C0}    & 0.044   & 0.044      \\
  \multirow{-1}{*}{Disability} & \multirow{-2}{*}{\cellcolor[HTML]{C0C0C0}0.726}    & \cellcolor[HTML]{343434}{\color[HTML]{FFFFFF} (-97.93\%)}   & (-96.56\%) & \multirow{-2}{*}{\cellcolor[HTML]{C0C0C0}0.303}    & (-92.74\%)   & (-92.74\%) & \multirow{-2}{*}{\cellcolor[HTML]{C0C0C0}0.294}    & \cellcolor[HTML]{343434}{\color[HTML]{FFFFFF} (-90.14\%)} & (-89.46\%) & \multirow{-2}{*}{\cellcolor[HTML]{C0C0C0}0.318} & (-86.16\%)   & (-86.16\%) \\\hline
 \multirow{-1}{*}{Std.P}    & \cellcolor[HTML]{343434}{\color[HTML]{FFFFFF} }    & \cellcolor[HTML]{C0C0C0}0.129    & 0.128      & \cellcolor[HTML]{C0C0C0}     & 0.076   & 0.076      & \cellcolor[HTML]{C0C0C0}     & 0.026   & 0.026      &    & \cellcolor[HTML]{343434}{\color[HTML]{FFFFFF} 0.048}      & 0.049      \\
  \multirow{-1}{*}{Gender}     & \multirow{-2}{*}{\cellcolor[HTML]{343434}{\color[HTML]{FFFFFF} 0.036}} & \cellcolor[HTML]{C0C0C0}(258.33\%)    & (255.56\%) & \multirow{-2}{*}{\cellcolor[HTML]{C0C0C0}0.079}    & (-3.8\%)   & (-3.8\%)   & \multirow{-2}{*}{\cellcolor[HTML]{C0C0C0}0.027}    & (-3.7\%)   & (-3.7\%)   & \multirow{-2}{*}{0.049}   & \cellcolor[HTML]{343434}{\color[HTML]{FFFFFF} (-2.04\%)}  & (0.\%)     \\\hline
 \multirow{-1}{*}{Std.P}   & \cellcolor[HTML]{C0C0C0}     & \cellcolor[HTML]{343434}{\color[HTML]{FFFFFF} 0.061}     & 0.062      & \cellcolor[HTML]{C0C0C0}     & 0.055   & 0.055      & \cellcolor[HTML]{C0C0C0}     & 0.027   & 0.027      & \cellcolor[HTML]{C0C0C0}    & 0.047   & 0.047      \\
  \multirow{-1}{*}{Health}     & \multirow{-2}{*}{\cellcolor[HTML]{C0C0C0}0.067}    & \cellcolor[HTML]{343434}{\color[HTML]{FFFFFF} (-8.96\%)} & (-7.46\%)  & \multirow{-2}{*}{\cellcolor[HTML]{C0C0C0}0.056}    & (-1.79\%)    & (-1.79\%)  & \multirow{-1}{*}{\cellcolor[HTML]{C0C0C0}0.028}    & (-3.57\%)    & (-3.57\%)  & \multirow{-2}{*}{\cellcolor[HTML]{C0C0C0}0.048} & (-2.08\%)    & (-2.08\%)  \\\hline
 \multirow{-1}{*}{Std.D}     & \cellcolor[HTML]{C0C0C0}     & \cellcolor[HTML]{343434}{\color[HTML]{FFFFFF} 0.261}     & 0.298      & \cellcolor[HTML]{C0C0C0}     & \cellcolor[HTML]{C0C0C0}0.108    & 0.107      & \cellcolor[HTML]{C0C0C0}     & \cellcolor[HTML]{343434}{\color[HTML]{FFFFFF} 0.118}      & 0.132      & \cellcolor[HTML]{C0C0C0}    & \cellcolor[HTML]{343434}{\color[HTML]{FFFFFF} 0.118}      & 0.160      \\
 \multirow{-1}{*}{Gender}     & \multirow{-2}{*}{\cellcolor[HTML]{C0C0C0}2.481}    & \cellcolor[HTML]{343434}{\color[HTML]{FFFFFF} (-89.48\%)}   & (-87.99\%) & \multirow{-2}{*}{\cellcolor[HTML]{C0C0C0}0.411}    & \cellcolor[HTML]{C0C0C0}(-73.72\%)    & (-73.97\%) & \multirow{-2}{*}{\cellcolor[HTML]{C0C0C0}0.185}    & \cellcolor[HTML]{343434}{\color[HTML]{FFFFFF} (-36.22\%)} & (-28.65\%) & \multirow{-2}{*}{\cellcolor[HTML]{C0C0C0}0.190} & \cellcolor[HTML]{343434}{\color[HTML]{FFFFFF} (-37.89\%)} & (-15.79\%) \\\hline
 \multirow{-1}{*}{Std.D}     & \cellcolor[HTML]{C0C0C0}     & \cellcolor[HTML]{343434}{\color[HTML]{FFFFFF} 0.141}     & 0.740      & \cellcolor[HTML]{C0C0C0}     & \cellcolor[HTML]{343434}{\color[HTML]{FFFFFF} 0.070}      & 0.240      & \cellcolor[HTML]{C0C0C0}     & \cellcolor[HTML]{343434}{\color[HTML]{FFFFFF} 0.082}      & 0.106      & \cellcolor[HTML]{C0C0C0}    & \cellcolor[HTML]{343434}{\color[HTML]{FFFFFF} 0.087}      & 0.126      \\
  \multirow{-2}{*}{Debtor}     & \multirow{-2}{*}{\cellcolor[HTML]{C0C0C0}0.816}    & \cellcolor[HTML]{343434}{\color[HTML]{FFFFFF} (-82.72\%)}   & (-9.31\%)  & \multirow{-2}{*}{\cellcolor[HTML]{C0C0C0}0.287}    & \cellcolor[HTML]{343434}{\color[HTML]{FFFFFF} (-75.61\%)} & (-16.38\%) & \multirow{-2}{*}{\cellcolor[HTML]{C0C0C0}0.139}    & \cellcolor[HTML]{343434}{\color[HTML]{FFFFFF} (-41.01\%)} & (-23.74\%) & \multirow{-2}{*}{\cellcolor[HTML]{C0C0C0}0.129} & \cellcolor[HTML]{343434}{\color[HTML]{FFFFFF} (-32.56\%)} & (-2.33\%)  \\\hline
\multirow{-1}{*}{\cellcolor[HTML]{FFFFFF}DNU}                            & \multicolumn{1}{r|}{\cellcolor[HTML]{000000}{\color[HTML]{FFFFFF} }}                        & \multicolumn{1}{r|}{\cellcolor[HTML]{000000}{\color[HTML]{FFFFFF} 0.012}}      & \multicolumn{1}{r|}{\cellcolor[HTML]{FFFFFF}{\color[HTML]{212121} 0.045}}      & \multicolumn{1}{r|}{\cellcolor[HTML]{000000}{\color[HTML]{FFFFFF} }}                        & \multicolumn{1}{r|}{\cellcolor[HTML]{000000}{\color[HTML]{FFFFFF} 0.075}}     & \multicolumn{1}{r|}{\cellcolor[HTML]{FFFFFF}{\color[HTML]{212121} 0.077}}     & \multicolumn{1}{r|}{\cellcolor[HTML]{D9D9D9}{\color[HTML]{212121} }}                        & \multicolumn{1}{r|}{\cellcolor[HTML]{D9D9D9}{\color[HTML]{212121} 0.222}}      & \multicolumn{1}{r|}{\cellcolor[HTML]{FFFFFF}{\color[HTML]{212121} 0.203}}      & \multicolumn{1}{r|}{\cellcolor[HTML]{000000}{\color[HTML]{FFFFFF} }}                        & \multicolumn{1}{r|}{\cellcolor[HTML]{000000}{\color[HTML]{FFFFFF} }}                        & \multicolumn{1}{r|}{\cellcolor[HTML]{FFFFFF}{\color[HTML]{212121} }}                        \\
  \multirow{-1}{*}{\cellcolor[HTML]{FFFFFF}Gender}     & \multicolumn{1}{r|}{\multirow{-2}{*}{\cellcolor[HTML]{000000}{\color[HTML]{FFFFFF} 0.036}}} & \multicolumn{1}{r|}{\cellcolor[HTML]{000000}{\color[HTML]{FFFFFF} (-66.67\%)}} & \multicolumn{1}{r|}{\cellcolor[HTML]{FFFFFF}{\color[HTML]{212121} (25.\%)}}    & \multicolumn{1}{r|}{\multirow{-2}{*}{\cellcolor[HTML]{000000}{\color[HTML]{FFFFFF} 0.049}}} & \multicolumn{1}{r|}{\cellcolor[HTML]{000000}{\color[HTML]{FFFFFF} (53.06\%)}} & \multicolumn{1}{r|}{\cellcolor[HTML]{FFFFFF}{\color[HTML]{212121} (57.14\%)}} & \multicolumn{1}{r|}{\multirow{-2}{*}{\cellcolor[HTML]{D9D9D9}{\color[HTML]{212121} 0.028}}} & \multicolumn{1}{r|}{\cellcolor[HTML]{D9D9D9}{\color[HTML]{212121} (692.86\%)}} & \multicolumn{1}{r|}{\cellcolor[HTML]{FFFFFF}{\color[HTML]{212121} (625.\%)}}   & \multicolumn{1}{r|}{\multirow{-2}{*}{\cellcolor[HTML]{000000}{\color[HTML]{FFFFFF} 0.000}}} & \multicolumn{1}{r|}{\multirow{-2}{*}{\cellcolor[HTML]{000000}{\color[HTML]{FFFFFF} 0.028}}} & \multicolumn{1}{r|}{\multirow{-2}{*}{\cellcolor[HTML]{FFFFFF}{\color[HTML]{212121} 0.039}}} \\\hline  
\multirow{-1}{*}{\cellcolor[HTML]{FFFFFF}DNU}                             & \multicolumn{1}{r|}{\cellcolor[HTML]{C0C0C0}{\color[HTML]{212121} }}                        & \multicolumn{1}{r|}{\cellcolor[HTML]{C0C0C0}{\color[HTML]{000000} 0.072}}      & \multicolumn{1}{r|}{\cellcolor[HTML]{FFFFFF}{\color[HTML]{212121} 0.046}}      & \multicolumn{1}{r|}{\cellcolor[HTML]{000000}{\color[HTML]{FFFFFF} }}                        & \multicolumn{1}{r|}{\cellcolor[HTML]{000000}{\color[HTML]{FFFFFF} 0.082}}     & \multicolumn{1}{r|}{\cellcolor[HTML]{FFFFFF}{\color[HTML]{212121} 0.085}}     & \multicolumn{1}{r|}{\cellcolor[HTML]{D9D9D9}{\color[HTML]{212121} }}                        & \multicolumn{1}{r|}{\cellcolor[HTML]{D9D9D9}{\color[HTML]{212121} 0.177}}      & \multicolumn{1}{r|}{\cellcolor[HTML]{FFFFFF}{\color[HTML]{212121} 0.133}}      & \multicolumn{1}{r|}{\cellcolor[HTML]{000000}{\color[HTML]{FFFFFF} }}                        & \multicolumn{1}{r|}{\cellcolor[HTML]{000000}{\color[HTML]{FFFFFF} }}                        & \multicolumn{1}{r|}{\cellcolor[HTML]{FFFFFF}{\color[HTML]{212121} }}                        \\
 \multirow{-1}{*}{\cellcolor[HTML]{FFFFFF}Age}        & \multicolumn{1}{r|}{\multirow{-2}{*}{\cellcolor[HTML]{C0C0C0}{\color[HTML]{212121} 0.082}}} & \multicolumn{1}{r|}{\cellcolor[HTML]{C0C0C0}{\color[HTML]{000000} (-85.37\%)}} & \multicolumn{1}{r|}{\cellcolor[HTML]{FFFFFF}{\color[HTML]{212121} (-43.9\%)}}  & \multicolumn{1}{r|}{\multirow{-2}{*}{\cellcolor[HTML]{000000}{\color[HTML]{FFFFFF} 0.078}}} & \multicolumn{1}{r|}{\cellcolor[HTML]{000000}{\color[HTML]{FFFFFF} (5.13\%)}}  & \multicolumn{1}{r|}{\cellcolor[HTML]{FFFFFF}{\color[HTML]{212121} (8.97\%)}}  & \multicolumn{1}{r|}{\multirow{-2}{*}{\cellcolor[HTML]{D9D9D9}{\color[HTML]{212121} 0.183}}} & \multicolumn{1}{r|}{\cellcolor[HTML]{D9D9D9}{\color[HTML]{212121} (-3.28\%)}}  & \multicolumn{1}{r|}{\cellcolor[HTML]{FFFFFF}{\color[HTML]{212121} (-27.32\%)}} & \multicolumn{1}{r|}{\multirow{-2}{*}{\cellcolor[HTML]{000000}{\color[HTML]{FFFFFF} 0.000}}} & \multicolumn{1}{r|}{\multirow{-2}{*}{\cellcolor[HTML]{000000}{\color[HTML]{FFFFFF} 0.000}}} & \multicolumn{1}{r|}{\multirow{-2}{*}{\cellcolor[HTML]{FFFFFF}{\color[HTML]{212121} 0.037}}} \\\hline  
\multirow{-1}{*}{\cellcolor[HTML]{FFFFFF}DNU}                             & \multicolumn{1}{r|}{\cellcolor[HTML]{C0C0C0}{\color[HTML]{212121} }}                        & \multicolumn{1}{r|}{\cellcolor[HTML]{C0C0C0}{\color[HTML]{212121} 0.040}}      & \multicolumn{1}{r|}{\cellcolor[HTML]{FFFFFF}{\color[HTML]{212121} 0.015}}      & \multicolumn{1}{r|}{\cellcolor[HTML]{000000}{\color[HTML]{FFFFFF} }}                        & \multicolumn{1}{r|}{\cellcolor[HTML]{000000}{\color[HTML]{FFFFFF} 0.050}}     & \multicolumn{1}{r|}{\cellcolor[HTML]{FFFFFF}{\color[HTML]{212121} 0.070}}     & \multicolumn{1}{r|}{\cellcolor[HTML]{D9D9D9}{\color[HTML]{212121} }}                        & \multicolumn{1}{r|}{\cellcolor[HTML]{D9D9D9}{\color[HTML]{212121} 0.165}}      & \multicolumn{1}{r|}{\cellcolor[HTML]{FFFFFF}{\color[HTML]{212121} 0.150}}      & \multicolumn{1}{r|}{\cellcolor[HTML]{000000}{\color[HTML]{FFFFFF} }}                        & \multicolumn{1}{r|}{\cellcolor[HTML]{C0C0C0}{\color[HTML]{212121} }}                        & \multicolumn{1}{r|}{\cellcolor[HTML]{FFFFFF}{\color[HTML]{212121} }}                        \\
  \multirow{-1}{*}{\cellcolor[HTML]{FFFFFF}Birthplace} & \multicolumn{1}{r|}{\multirow{-2}{*}{\cellcolor[HTML]{C0C0C0}{\color[HTML]{212121} 0.047}}} & \multicolumn{1}{r|}{\cellcolor[HTML]{C0C0C0}{\color[HTML]{212121} (-14.89\%)}} & \multicolumn{1}{r|}{\cellcolor[HTML]{FFFFFF}{\color[HTML]{212121} (-68.09\%)}} & \multicolumn{1}{r|}{\multirow{-2}{*}{\cellcolor[HTML]{000000}{\color[HTML]{FFFFFF} 0.051}}} & \multicolumn{1}{r|}{\cellcolor[HTML]{000000}{\color[HTML]{FFFFFF} (-1.96\%)}} & \multicolumn{1}{r|}{\cellcolor[HTML]{FFFFFF}{\color[HTML]{212121} (37.25\%)}} & \multicolumn{1}{r|}{\multirow{-2}{*}{\cellcolor[HTML]{D9D9D9}{\color[HTML]{212121} 0.164}}} & \multicolumn{1}{r|}{\cellcolor[HTML]{D9D9D9}{\color[HTML]{212121} (0.61\%)}}   & \multicolumn{1}{r|}{\cellcolor[HTML]{FFFFFF}{\color[HTML]{212121} (-8.54\%)}}  & \multicolumn{1}{r|}{\multirow{-2}{*}{\cellcolor[HTML]{000000}{\color[HTML]{FFFFFF} 0.000}}} & \multicolumn{1}{r|}{\multirow{-2}{*}{\cellcolor[HTML]{C0C0C0}{\color[HTML]{212121} 0.041}}} & \multicolumn{1}{r|}{\multirow{-2}{*}{\cellcolor[HTML]{FFFFFF}{\color[HTML]{212121} 0.040}}} \\
\hline  
\textbf{W/T/L} & \multicolumn{3}{c|}{8/0/7}  & \multicolumn{3}{c|}{9/0/6}& \multicolumn{3}{c|}{10/0/5} & \multicolumn{3}{c|}{8/0/7}  \\\hline
\end{tabular}}
\end{small} 
\end{center}
\end{table}
%\end{landscape}
\begin{figure}[htbp]
\centerline{\includegraphics[width=1\textwidth]{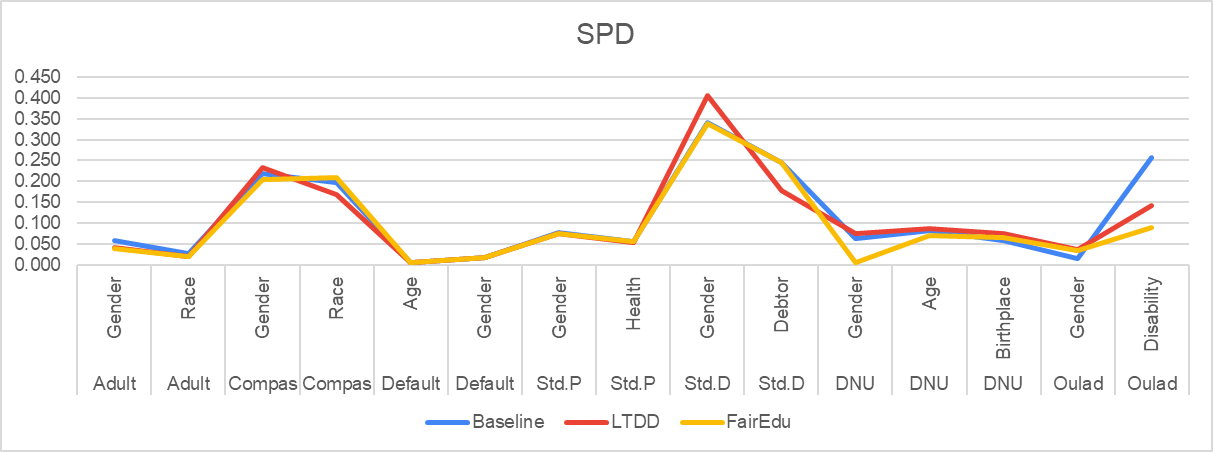}}
\caption{Comparison of SPD across methods}
\label{SPD_RF}
\end{figure}

% \begin{figure}[htbp]
%     \centering
%     \begin{subfigure}[b]{0.3\textwidth}
%         \includegraphics[width=\textwidth]{Figure/Change of |1-DI|_LR.png}
%         \caption{\(|1-DI|\) in Logistic Regression}
%         \label{Fig.5}
%     \end{subfigure}
%     \hfill
%     \begin{subfigure}[b]{0.3\textwidth}
%         \includegraphics[width=\textwidth]{Figure/Change of |1-DI|_RF.png}
%         \caption{\(|1-DI|\) in Random Forest}
%         \label{Fig.6}
%     \end{subfigure}
%     \hfill
%     \begin{subfigure}[b]{0.3\textwidth}
%         \includegraphics[width=\textwidth]{Figure/Change of 1-DI_DT.png}
%         \caption{\(|1-DI|\) in Decision Tree}
%         \label{Fig.7}
%     \end{subfigure}
%     \caption{Comparison of \(|1-DI|\) across different models}
%     \label{fig:comparison3}
% \end{figure}
A deeper comparison between FAIREDU and LTDD can be seen in Table \ref{tab: Compare Bef-Af Logistic Regression}. For the Logistic Regression model, we applied both LTDD and FAIREDU to all seven datasets, evaluating a total of 15 sensitive features. Each scenario was run 100 times to obtain average results, ensuring statistical significance and minimizing the impact of random fluctuations. We use the colored boxes to highlight the results (better or worse than the baseline). We also report the difference in percent change.  In total, across 60 fairness comparisons (win/tie/loss), FAIREDU achieved 35 wins and 25 losses against LTDD. These results demonstrate that FAIREDU provides superior performance in most situations when compared to LTDD.

\begin{figure}[htbp]
\centerline{\includegraphics[width=1\textwidth]{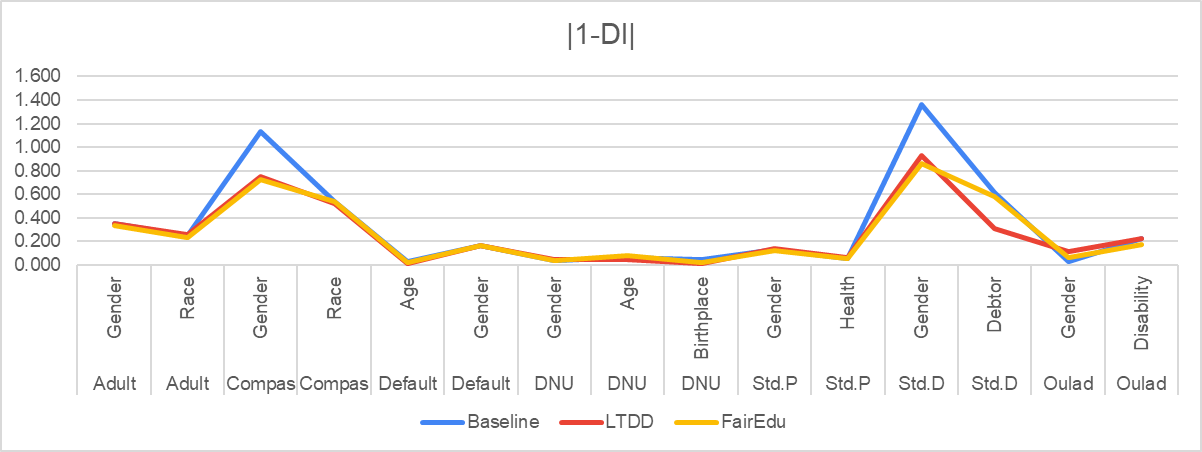}}
\caption{Comparison of $|1-DI|$ across methods}
\label{1-DI_DT}
\end{figure}

Besides that, the results of applying the Random Forest (RF) model, include without intervention and with the Fairedu or LTDD interventions, are summarized in Figure \ref{SPD_RF}. In this figure, color lines indicate fairness metric outcomes, where the lower line represents more improved fairness. As shown in Figure \ref{SPD_RF}, the Fairedu intervention either outperformed or matched the performance of LTDD and the original model in most cases. Specifically, Fairedu surpass LTDD in 11 out of 15 cases based on the SPD measure, which covers the majority of scenarios tested across 7 datasets and 15 sensitive features listed in Table \ref{Table Dataset}. 

Similarly, for the Decision Tree (DT) model, we experiments conducted using the same 7 datasets, which are mentioned in Table \ref{Table Dataset}, in cases without intervention, as well as with Fairedu and LTDD interventions. The results of the fairness metrics $|1-DI|$ demonstrate that Fairedu outperformed LTDD in 9 out of 15 cases, which results are summaried in Figure \ref{1-DI_DT}, again constituting the majority of experimental scenarios. 

In summary, across all three models: Logistic Regression (LT), Random Forest (RF), and Decision Tree (DT) the Fairedu method consistently yields positive results, showing superiority over the previous LTDD method in most cases.
\begin{note}[label={note:3}]{The Fairness of FAIREDU and state-of-the-art methods}
FAIREDU has demonstrated superior fairness compared to other fairness enhancement methods in two key areas:
\begin{itemize}
        \item Simultaneously addressing and improving fairness across multiple sensitive features within the dataset.
        \item FAIREDU has improved the equity indicators after the intervention, specifically:
    \begin{itemize}
        \item $|1-DI|$ reduced up to 96.7$\%$ (wrt. Gender in Oulad set)
        \item SPD reduced up to 88.55$\%$ (wrt. Gender in Adult set)
        \item AOD reduced up to 85.79$\%$ (wrt. Race in Compas set)
        \item EOD reduced up to 84.87$\%$ (wrt. Gender in Oulad set)
    \end{itemize}
 \end{itemize}
\end{note}

\subsection{RQ4 - How effectively does FAIREDU balance fairness and model performance relative to state-of-the-art methods?}
\label{subsec: RQ4}

To evaluate how well FAIREDU balances fairness and model performance, we conducted statistical analyses on three models: Logistic Regression, Random Forest, and Decision Tree, applied across seven datasets and seven sensitive features. The comparison was made between three methods: no intervention (Origin), an intervention using LTDD, and the fairness intervention using FAIREDU, as shown in Table \ref{tab: Compare ACC_Recall}.

\textbf{Model Performance Impact}: As shown in Table \ref{tab: Compare ACC_Recall}, FAIREDU's intervention aimed to improve fairness across all sensitive features, and the results indicate that overall model performance did not significantly decline. For accuracy (ACC), FAIREDU outperformed both Origin and LTDD in 9 out of 45 cases and tied in 13 cases. While it underperformed in 23 cases, the performance reduction was minimal, with the highest deviation being 5.71$\%$ in the Decision Tree model on the DNU-BP dataset. In this case, the accuracy decreased by no more than 0.056, highlighting that the performance drop was relatively small. For recall, FAIREDU outperformed the other methods in 4 cases, tied in 15 cases, and underperformed in 26 cases. Similar to accuracy, the deviations were not significant, with the largest reduction being 9.6$\%$ in the Logistic Regression model on the Adult dataset, where recall decreased by 0.096 at most.

\textbf{Fairness-Performance Tradeoff}: Despite these performance fluctuations, the application of FAIREDU demonstrated its effectiveness in improving fairness while minimally affecting model accuracy and recall. The small deviations in performance suggest that FAIREDU manages to maintain a balance between fairness and model effectiveness, which is crucial when implementing fairness interventions in practical applications. While fairness-focused interventions can sometimes lead to significant reductions in performance, FAIREDU shows that it is possible to enhance fairness with minimal compromises.

%To assess the impact of fairness interventions using the FAIREDU method on model performance, we conducted a statistical analysis of the Logistic Regression, Random Forest, and Decision Tree models across seven datasets and seven sensitive features mentioned in Section \ref{subsec: Data}. The analysis compared three methods: no intervention (Origin), intervention with LTDD (LTDD), and intervention with FAIREDU (FAIREDU). As shown in Table \ref{tab: Compare ACC_Recall}, after applying FAIREDU to improve fairness for all sensitive features, the overall model performance did not decrease significantly. In some instances, model performance even improved.

\begin{table}[]
\begin{small}
\begin{center}
\caption{Performance measure before and after EDUFAIR}
\label{tab: Compare ACC_Recall}
\scalebox{0.7}{ % Giảm kích thước bảng xuống 90%
\begin{tabular}
{|p{1cm}|p{2.2cm}|p{1cm}|p{3cm}|p{3cm}|p{1cm}|p{3cm}|p{3cm}|} \hline 
\multicolumn{1}{|c|}{} & \multicolumn{1}{c|}{\textbf{Indicators}}   & \multicolumn{3}{c|}{\textbf{ACC}}& \multicolumn{3}{c|}{\textbf{Recall}}   \\ \cline{2-8} 
\multicolumn{1}{|c|}{\multirow{-2}{*}{\textbf{Model}}} & \multicolumn{1}{c|}{\textit{\textbf{Methods}}} & \multicolumn{1}{c|}{\textit{\textbf{Before}}}  & \multicolumn{1}{c|}{\textit{\textbf{\begin{tabular}[c]{@{}c@{}}LTDD\_af   \\  (\%change)\end{tabular}}}} & \multicolumn{1}{c|}{\textit{\textbf{\begin{tabular}[c]{@{}c@{}}FAIREDU\_af\\  (\%change)\end{tabular}}}} & \multicolumn{1}{c|}{\textit{\textbf{Before}}}  & \multicolumn{1}{c|}{\textit{\textbf{\begin{tabular}[c]{@{}c@{}}LTDD\_af \\  (\%change)\end{tabular}}}} & \multicolumn{1}{c|}{\textit{\textbf{\begin{tabular}[c]{@{}c@{}}FAIREDU\_af\\  (\%change)\end{tabular}}}} \\ \hline
\multicolumn{1}{|l|}{LR}  & A-Gen  & \multicolumn{1}{l|}{0.821} & \multicolumn{1}{r|}{\textbf{0.806(-1.5\%)}}  & \textbf{0.803(-2.19\%)}  & \multicolumn{1}{r|}{\textbf{0.417}}& \multicolumn{1}{r|}{\textbf{0.34(-7.7\%)}} & \textbf{0.321(-9.6\%)}   \\ \hline
\multicolumn{1}{|l|}{RF}& A-Gen  & \multicolumn{1}{l|}{0.816} & \multicolumn{1}{r|}{\textbf{0.813(-0.3\%)}}  & \textbf{0.812(-0.4\%)}   & \multicolumn{1}{r|}{\textbf{0.246}}& \multicolumn{1}{r|}{\textbf{0.237(-0.9\%)}}& \textbf{0.234(-1.2\%)}   \\ \hline
\multicolumn{1}{|l|}{DT}& A-Gen  & \multicolumn{1}{l|}{0.806} & \multicolumn{1}{r|}{\textbf{0.805(-0.1\%)}}  & \textbf{0.801(-0.5\%)}   & \multicolumn{1}{r|}{\textbf{0.525}}& \multicolumn{1}{r|}{\textbf{0.522(-0.3\%)}}& \textbf{0.508(-1.7\%)}   \\ \hline
\multicolumn{1}{|l|}{LR}  & A-race & \multicolumn{1}{l|}{0.821} & \multicolumn{1}{r|}{\textbf{0.821(0.\%)}}& \textbf{0.803(-1.8\%)}   & \multicolumn{1}{r|}{\textbf{0.417}}& \multicolumn{1}{r|}{\textbf{0.42(0.3\%)}}  & \textbf{0.321(-9.6\%)}   \\ \hline
\multicolumn{1}{|l|}{RF}& A-race & \multicolumn{1}{l|}{0.816} & \multicolumn{1}{r|}{\textbf{0.813(-0.3\%)}}  & \textbf{0.812(-0.4\%)}   & \multicolumn{1}{r|}{\textbf{0.246}}& \multicolumn{1}{r|}{\textbf{0.232(-1.4\%)}}& \cellcolor[HTML]{D9D9D9}\textbf{0.234(-1.2\%)}   \\ \hline
\multicolumn{1}{|l|}{DT}& A-race & \multicolumn{1}{l|}{0.806} & \multicolumn{1}{r|}{\textbf{0.806(0.\%)}}& \textbf{0.801(-0.5\%)}   & \multicolumn{1}{r|}{\textbf{0.525}}& \multicolumn{1}{r|}{\textbf{0.524(-0.1\%)}}& \textbf{0.508(-1.7\%)}   \\ \hline
\multicolumn{1}{|l|}{LR}  & C-Gen & \multicolumn{1}{r|}{\textbf{0.641}}& \multicolumn{1}{r|}{\textbf{0.64(-0.1\%)}}   & \cellcolor[HTML]{000000}{\color[HTML]{FFFFFF} \textbf{0.646(0.5\%)}} & \multicolumn{1}{r|}{\textbf{0.566}}& \multicolumn{1}{r|}{\textbf{0.592(2.6\%)}} & \cellcolor[HTML]{D9D9D9}\textbf{0.568(0.2\%)}\\ \hline
\multicolumn{1}{|l|}{RF}& C-Gen & \multicolumn{1}{r|}{\textbf{0.658}}& \multicolumn{1}{r|}{\textbf{0.662(0.4\%)}}   & \textbf{0.658(0.\%)} & \multicolumn{1}{r|}{\textbf{0.507}}& \multicolumn{1}{r|}{\textbf{0.542(3.5\%)}} & \cellcolor[HTML]{D9D9D9}\textbf{0.508(0.1\%)}\\ \hline
\multicolumn{1}{|l|}{DT}& C-Gen & \multicolumn{1}{l|}{0.662} & \multicolumn{1}{r|}{\textbf{0.664(0.2\%)}}   & \cellcolor[HTML]{D9D9D9}\textbf{0.663(0.1\%)}& \multicolumn{1}{r|}{\textbf{0.585}}& \multicolumn{1}{r|}{\textbf{0.577(-0.8\%)}}& \textbf{0.576(-0.9\%)}   \\ \hline
\multicolumn{1}{|l|}{LR}  & C-race& \multicolumn{1}{r|}{\textbf{0.641}}& \multicolumn{1}{r|}{\textbf{0.642(0.1\%)}}   & \cellcolor[HTML]{000000}{\color[HTML]{FFFFFF} \textbf{0.646(0.5\%)}} & \multicolumn{1}{r|}{\textbf{0.566}}& \multicolumn{1}{r|}{\textbf{0.57(0.4\%)}}  & \cellcolor[HTML]{D9D9D9}\textbf{0.568(0.2\%)}\\ \hline
\multicolumn{1}{|l|}{RF}& C-race& \multicolumn{1}{r|}{\textbf{0.658}}& \multicolumn{1}{r|}{\textbf{0.664(0.6\%)}}   & \cellcolor[HTML]{D9D9D9}\textbf{0.658(0.\%)} & \multicolumn{1}{r|}{\textbf{0.507}}& \multicolumn{1}{r|}{\textbf{0.552(4.5\%)}} & \cellcolor[HTML]{D9D9D9}\textbf{0.508(0.1\%)}\\ \hline
\multicolumn{1}{|l|}{DT}& C-race& \multicolumn{1}{l|}{0.662} & \multicolumn{1}{r|}{\textbf{0.661(-0.1\%)}}  & \cellcolor[HTML]{000000}{\color[HTML]{FFFFFF} \textbf{0.663(0.1\%)}} & \multicolumn{1}{r|}{\textbf{0.585}}& \multicolumn{1}{r|}{\textbf{0.583(-0.2\%)}}& \textbf{0.576(-0.9\%)}   \\ \hline
\multicolumn{1}{|l|}{LR}  & D-age& \multicolumn{1}{r|}{\textbf{0.81}} & \multicolumn{1}{r|}{\textbf{0.81(0.\%)}} & \textbf{0.809(-0.1\%)}   & \multicolumn{1}{r|}{\textbf{0.23}} & \multicolumn{1}{r|}{\textbf{0.229(-0.1\%)}}& \textbf{0.228(-0.2\%)}   \\ \hline
\multicolumn{1}{|l|}{RF}& D-age& \multicolumn{1}{r|}{\textbf{0.808}}& \multicolumn{1}{r|}{\textbf{0.81(0.2\%)}}& \cellcolor[HTML]{D9D9D9}\textbf{0.808(0.\%)} & \multicolumn{1}{r|}{\textbf{0.233}}& \multicolumn{1}{r|}{\textbf{0.249(1.6\%)}} & \cellcolor[HTML]{D9D9D9}\textbf{0.236(0.3\%)}\\ \hline
\multicolumn{1}{|l|}{DT}& D-age& \multicolumn{1}{l|}{0.821} & \multicolumn{1}{r|}{\textbf{0.821(0.\%)}}& \cellcolor[HTML]{D9D9D9}\textbf{0.821(0.\%)} & \multicolumn{1}{r|}{\textbf{0.37}} & \multicolumn{1}{r|}{\textbf{0.368(-0.2\%)}}& \textbf{0.365(-0.5\%)}   \\ \hline
\multicolumn{1}{|l|}{LR}  & D-Gen& \multicolumn{1}{r|}{\textbf{0.81}} & \multicolumn{1}{r|}{\textbf{0.809(-0.1\%)}}  & \cellcolor[HTML]{D9D9D9}\textbf{0.809(-0.1\%)}   & \multicolumn{1}{r|}{\textbf{0.23}} & \multicolumn{1}{r|}{\textbf{0.227(-0.3\%)}}& \cellcolor[HTML]{D9D9D9}\textbf{0.228(-0.2\%)}   \\ \hline
\multicolumn{1}{|l|}{RF}& D-Gen& \multicolumn{1}{r|}{\textbf{0.808}}& \multicolumn{1}{r|}{\textbf{0.81(0.2\%)}}& \cellcolor[HTML]{D9D9D9}\textbf{0.808(0.\%)} & \multicolumn{1}{r|}{\textbf{0.233}}& \multicolumn{1}{r|}{\textbf{0.251(1.8\%)}} & \cellcolor[HTML]{D9D9D9}\textbf{0.236(0.3\%)}\\ \hline
\multicolumn{1}{|l|}{DT}& D-Gen& \multicolumn{1}{l|}{0.821} & \multicolumn{1}{r|}{\textbf{0.82(-0.1\%)}}   & \cellcolor[HTML]{000000}{\color[HTML]{FFFFFF} \textbf{0.821(0.\%)}}  & \multicolumn{1}{r|}{\textbf{0.37}} & \multicolumn{1}{r|}{\textbf{0.368(-0.2\%)}}& \textbf{0.365(-0.5\%)}   \\ \hline
\multicolumn{1}{|l|}{LR}  & O-Gen   & \multicolumn{1}{r|}{\textbf{0.588}}& \multicolumn{1}{r|}{\textbf{0.585(-0.3\%)}}  & \textbf{0.582(-0.6\%)}   & \multicolumn{1}{r|}{\textbf{0.473}}& \multicolumn{1}{r|}{\textbf{0.468(-0.5\%)}} & \textbf{0.453(-2.\%)}\\ \hline
\multicolumn{1}{|l|}{RF} & O-Gen   & \multicolumn{1}{r|}{\textbf{0.58}} & \multicolumn{1}{r|}{\textbf{0.581(0.1\%)}}   & \cellcolor[HTML]{D9D9D9}\textbf{0.58(0.\%)}  & \multicolumn{1}{r|}{\textbf{0.482}}& \multicolumn{1}{r|}{\textbf{0.465(-1.7\%)}}& \cellcolor[HTML]{D9D9D9}\textbf{0.478(-0.4\%)}   \\ \hline
\multicolumn{1}{|l|}{DT} & O-Gen   & \multicolumn{1}{l|}{0.578} & \multicolumn{1}{r|}{\textbf{0.579(0.1\%)}}   & \cellcolor[HTML]{D9D9D9}\textbf{0.578(0.\%)} & \multicolumn{1}{r|}{\textbf{0.518}}& \multicolumn{1}{r|}{\textbf{0.479(-3.9\%)}}& \cellcolor[HTML]{D9D9D9}\textbf{0.492(-2.6\%)}   \\ \hline
\multicolumn{1}{|l|}{LR}  & O-disability   & \multicolumn{1}{r|}{\textbf{0.588}}& \multicolumn{1}{r|}{\textbf{0.583(-0.5\%)}}  & \textbf{0.582(-0.6\%)}   & \multicolumn{1}{r|}{\textbf{0.473}}& \multicolumn{1}{r|}{\textbf{0.458(-1.5\%)}}& \textbf{0.453(-2.\%)}\\ \hline
\multicolumn{1}{|l|}{RF}& O-disability   & \multicolumn{1}{r|}{\textbf{0.58}} & \multicolumn{1}{r|}{\textbf{0.579(-0.1\%)}}  & \cellcolor[HTML]{000000}{\color[HTML]{FFFFFF} \textbf{0.58(0.\%)}}   & \multicolumn{1}{r|}{\textbf{0.482}}& \multicolumn{1}{r|}{\textbf{0.483(0.1\%)}} & \textbf{0.478(-0.4\%)}   \\ \hline
\multicolumn{1}{|l|}{DT}& O-disability   & \multicolumn{1}{l|}{0.578} & \multicolumn{1}{r|}{\textbf{0.578(0.\%)}}& \cellcolor[HTML]{D9D9D9}\textbf{0.578(0.\%)} & \multicolumn{1}{r|}{\textbf{0.518}}& \multicolumn{1}{r|}{\textbf{0.518(0.\%)}}  & \textbf{0.492(-2.6\%)}   \\ \hline
\multicolumn{1}{|l|}{LR}  & S-P-Gen   & \multicolumn{1}{r|}{\textbf{0.935}}& \multicolumn{1}{r|}{\textbf{0.935(0.\%)}}& \textbf{0.936(0.1\%)}& \multicolumn{1}{r|}{\textbf{0.913}}& \multicolumn{1}{r|}{\textbf{0.912(-0.1\%)}}& \cellcolor[HTML]{000000}{\color[HTML]{FFFFFF} \textbf{0.913(0.\%)}}  \\ \hline
\multicolumn{1}{|l|}{RF}& S-P-Gen   & \multicolumn{1}{r|}{\textbf{0.93}} & \multicolumn{1}{r|}{\textbf{0.938(0.8\%)}}   & \textbf{0.935(0.5\%)}& \multicolumn{1}{r|}{\textbf{0.914}}& \multicolumn{1}{r|}{\textbf{0.912(-0.2\%)}}& \textbf{0.909(-0.5\%)}   \\ \hline
\multicolumn{1}{|l|}{DT}& S-P-Gen   & \multicolumn{1}{l|}{0.932} & \multicolumn{1}{r|}{\textbf{0.932(0.\%)}}& \textbf{0.928(-0.4\%)}   & \multicolumn{1}{r|}{\textbf{0.908}}& \multicolumn{1}{r|}{\textbf{0.908(0.\%)}}  & \cellcolor[HTML]{000000}{\color[HTML]{FFFFFF} \textbf{0.91(0.2\%)}}  \\ \hline
\multicolumn{1}{|l|}{LR}  & S-P-health& \multicolumn{1}{r|}{\textbf{0.935}}& \multicolumn{1}{r|}{\textbf{0.935(0.\%)}}& \cellcolor[HTML]{000000}{\color[HTML]{FFFFFF} \textbf{0.936(0.1\%)}} & \multicolumn{1}{r|}{\textbf{0.913}} & \multicolumn{1}{r|}{\textbf{0.913(0.\%)}}  & \cellcolor[HTML]{D9D9D9}\textbf{0.913(0.\%)} \\ \hline
\multicolumn{1}{|l|}{RF}& S-P-health & \multicolumn{1}{r|}{\textbf{0.93}} & \multicolumn{1}{r|}{\textbf{0.937(0.7\%)}}   & \textbf{0.935(0.5\%)}& \multicolumn{1}{r|}{\textbf{0.914}}& \multicolumn{1}{r|}{\textbf{0.911(-0.3\%)}}& \textbf{0.909(-0.5\%)}   \\ \hline
\multicolumn{1}{|l|}{DT}& S-P-health & \multicolumn{1}{l|}{0.932} & \multicolumn{1}{r|}{\textbf{0.931(-0.1\%)}}  & \textbf{0.928(-0.4\%)}   & \multicolumn{1}{r|}{\textbf{0.908}}& \multicolumn{1}{r|}{\textbf{0.907(-0.1\%)}}& \cellcolor[HTML]{000000}{\color[HTML]{FFFFFF} \textbf{0.91(0.2\%)}}  \\ \hline
\multicolumn{1}{|l|}{LR}   & S-D-Deb & \multicolumn{1}{r|}{\textbf{0.843}} & \multicolumn{1}{r|}{\textbf{0.821(-2.2\%)}}  & \textbf{0.819(-2.4\%)}   & \multicolumn{1}{r|}{\textbf{0.885}}& \multicolumn{1}{r|}{\textbf{0.859(-2.6\%)}}& \textbf{0.805(-8.\%)}\\ \hline
\multicolumn{1}{|l|}{RF}   & S-D-Deb & \multicolumn{1}{r|}{\textbf{0.827}}& \multicolumn{1}{r|}{\textbf{0.827(0.\%)}}& \textbf{0.825(-0.2\%)}   & \multicolumn{1}{r|}{\textbf{0.88}} & \multicolumn{1}{r|}{\textbf{0.885(0.5\%)}} & \textbf{0.878(-0.2\%)}   \\ \hline
\multicolumn{1}{|l|}{DT}   & S-D-Deb & \multicolumn{1}{l|}{0.819} & \multicolumn{1}{r|}{\textbf{0.818(-0.1\%)}}  & \textbf{0.817(-0.2\%)}   & \multicolumn{1}{r|}{\textbf{0.869}}& \multicolumn{1}{r|}{\textbf{0.883(1.4\%)}} & \cellcolor[HTML]{000000}{\color[HTML]{FFFFFF} \textbf{0.886(1.7\%)}} \\ \hline
\multicolumn{1}{|l|}{LR}   & S-D-Gen & \multicolumn{1}{r|}{\textbf{0.843}}& \multicolumn{1}{r|}{\textbf{0.827(-1.6\%)}}  & \textbf{0.819(-2.4\%)}   & \multicolumn{1}{r|}{\textbf{0.885}}& \multicolumn{1}{r|}{\textbf{0.876(-0.9\%)}}& \textbf{0.805(-8.\%)}\\ \hline
\multicolumn{1}{|l|}{RF}   & S-D-Gen & \multicolumn{1}{r|}{\textbf{0.827}}& \multicolumn{1}{r|}{\textbf{0.824(-0.3\%)}}  & \cellcolor[HTML]{D9D9D9}\textbf{0.825(-0.2\%)}   & \multicolumn{1}{r|}{\textbf{0.88}} & \multicolumn{1}{r|}{\textbf{0.897(1.7\%)}} & \textbf{0.878(-0.2\%)}   \\ \hline
\multicolumn{1}{|l|}{DT}   & S-D-Gen & \multicolumn{1}{l|}{0.819} & \multicolumn{1}{r|}{\textbf{0.812(-0.7\%)}}  & \cellcolor[HTML]{D9D9D9}\textbf{0.817(-0.2\%)}   & \multicolumn{1}{r|}{\textbf{0.869}}& \multicolumn{1}{r|}{\textbf{0.903(3.4\%)}} & \cellcolor[HTML]{D9D9D9}\textbf{0.886(1.7\%)}\\ \hline
\multicolumn{1}{|l|}{LR}   & S-DNU-Gen   & \multicolumn{1}{r|}{\cellcolor[HTML]{FFFFFF}{\color[HTML]{212121} \textbf{0.907}}} & \multicolumn{1}{r|}{{\color[HTML]{212121} \textbf{0.917(1.1\%)}}}& \cellcolor[HTML]{000000}{\color[HTML]{FFFFFF} \textbf{0.934(2.98\%)}}& \multicolumn{1}{r|}{{\color[HTML]{212121} \textbf{1}}} & \multicolumn{1}{r|}{{\color[HTML]{212121} \textbf{0.996(-0.4\%)}}} & {\color[HTML]{212121} \textbf{0.969(-3.1\%)}}\\ \hline
\multicolumn{1}{|l|}{RF}   & S-DNU-Gen   & \multicolumn{1}{r|}{\cellcolor[HTML]{FFFFFF}{\color[HTML]{212121} \textbf{0.941}}} & \multicolumn{1}{r|}{{\color[HTML]{212121} \textbf{0.93(-1.17\%)}}}   & \cellcolor[HTML]{D9D9D9}{\color[HTML]{212121} \textbf{0.932(-0.96\%)}}   & \multicolumn{1}{r|}{{\color[HTML]{212121} \textbf{1}}} & \multicolumn{1}{r|}{{\color[HTML]{212121} \textbf{0.999(-0.1\%)}}} & {\color[HTML]{212121} \textbf{0.99(-1.\%)}}  \\ \hline
\multicolumn{1}{|l|}{DT}   & S-DNU-Gen   & \multicolumn{1}{l|}{0.93}  & \multicolumn{1}{r|}{{\color[HTML]{212121} \textbf{0.925(-0.54\%)}}}  & {\color[HTML]{212121} \textbf{0.891(-4.19\%)}}   & \multicolumn{1}{r|}{{\color[HTML]{212121} \textbf{0.978}}} & \multicolumn{1}{r|}{{\color[HTML]{212121} \textbf{0.928(-5.11\%)}}}& \cellcolor[HTML]{D9D9D9}{\color[HTML]{212121} \textbf{0.928(-5.11\%)}}   \\ \hline
\multicolumn{1}{|l|}{LR}   & S-DNU-Age  & \multicolumn{1}{r|}{\cellcolor[HTML]{FFFFFF}{\color[HTML]{212121} \textbf{0.91}}}  & \multicolumn{1}{r|}{{\color[HTML]{212121} \textbf{0.925(1.65\%)}}}   & \cellcolor[HTML]{000000}{\color[HTML]{FFFFFF} \textbf{0.934(2.64\%)}}& \multicolumn{1}{r|}{{\color[HTML]{212121} \textbf{1}}} & \multicolumn{1}{r|}{{\color[HTML]{212121} \textbf{0.969(-3.1\%)}}} & \cellcolor[HTML]{D9D9D9}{\color[HTML]{212121} \textbf{0.969(-3.1\%)}}\\ \hline
\multicolumn{1}{|l|}{RF}   & S-DNU-Age  & \multicolumn{1}{r|}{\cellcolor[HTML]{FFFFFF}{\color[HTML]{212121} \textbf{0.938}}} & \multicolumn{1}{r|}{{\color[HTML]{212121} \textbf{0.932(-0.64\%)}}}  & \cellcolor[HTML]{D9D9D9}{\color[HTML]{212121} \textbf{0.932(-0.64\%)}}   & \multicolumn{1}{r|}{{\color[HTML]{212121} \textbf{1}}} & \multicolumn{1}{r|}{{\color[HTML]{212121} \textbf{0.99(-1.\%)}}}   & \cellcolor[HTML]{D9D9D9}{\color[HTML]{212121} \textbf{0.99(-1.\%)}}  \\ \hline
\multicolumn{1}{|l|}{DT}   & S-DNU-Age  & \multicolumn{1}{l|}{0.942} & \multicolumn{1}{r|}{{\color[HTML]{212121} \textbf{0.934(-0.85\%)}}}  & {\color[HTML]{212121} \textbf{0.891(-5.41\%)}}   & \multicolumn{1}{r|}{{\color[HTML]{212121} \textbf{0.977}}} & \multicolumn{1}{r|}{{\color[HTML]{212121} \textbf{0.973(-0.41\%)}}}& {\color[HTML]{212121} \textbf{0.928(-5.02\%)}}   \\ \hline
\multicolumn{1}{|l|}{LR}   & S-DNU-BP   & \multicolumn{1}{r|}{\cellcolor[HTML]{FFFFFF}{\color[HTML]{212121} \textbf{0.908}}} & \multicolumn{1}{r|}{{\color[HTML]{212121} \textbf{0.912(0.44\%)}}}   & \cellcolor[HTML]{000000}{\color[HTML]{FFFFFF} \textbf{0.934(2.86\%)}}& \multicolumn{1}{r|}{{\color[HTML]{212121} \textbf{1}}} & \multicolumn{1}{r|}{{\color[HTML]{212121} \textbf{1.(0. \%)}}}  & {\color[HTML]{212121} \textbf{0.969(-3.1\%)}}\\ \hline
\multicolumn{1}{|l|}{RF}   & S-DNU-BP   & \multicolumn{1}{r|}{\cellcolor[HTML]{FFFFFF}{\color[HTML]{212121} \textbf{0.941}}} & \multicolumn{1}{r|}{{\color[HTML]{212121} \textbf{0.94(-0.11\%)}}}   & {\color[HTML]{212121} \textbf{0.932(-0.96\%)}}   & \multicolumn{1}{r|}{{\color[HTML]{212121} \textbf{1}}} & \multicolumn{1}{r|}{{\color[HTML]{212121} \textbf{1.(0.\%)}}}  & {\color[HTML]{212121} \textbf{0.99(-1.\%)}}  \\ \hline
\multicolumn{1}{|l|}{DT}   & S-DNU-BP   & \multicolumn{1}{l|}{0.945} & \multicolumn{1}{r|}{{\color[HTML]{212121} \textbf{0.947(0.21\%)}}}   & {\color[HTML]{212121} \textbf{0.891(-5.71\%)}}   & \multicolumn{1}{r|}{{\color[HTML]{212121} \textbf{0.981}}} & \multicolumn{1}{r|}{{\color[HTML]{212121} \textbf{0.982(0.1\%)}}}  & {\color[HTML]{212121} \textbf{0.928(-5.4\%)}}\\ \hline
\multicolumn{2}{|c|}{\textbf{W/T/L}} & \multicolumn{3}{c|}{\textbf{9/13/23}}& \multicolumn{3}{c|}{\textbf{4/15/26}}  \\ \hline

\end{tabular}}
\end{center}
\end{small}
\end{table}

%To emphasize these results, we used color coding: black indicates superior performance (higher accuracy), and gray signifies performance that is equal to at least one of the other two methods.
%The analysis reveals that for accuracy (ACC), FAIREDU outperformed both the Origin and LTDD methods in 9 out of 45 cases, and tied in 13 cases. Although it underperformed in 23 cases, the performance reduction was minimal, with the highest deviation being 5.71$\%$ in the Decision Tree model for the S$\_$DNU-BP dataset (ACC decreased by no more than 0.056). For recall, FAIREDU outperformed both Origin and LTDD in 4 cases, tied in 15, and underperformed in 26 cases. Similar to accuracy, the deviations were not significant, with the highest deviation being 9.6$\%$ in the Logistic Regression model on the Adult dataset (recall decreased by no more than 0.096). These results demonstrate that the application of FAIREDU does not significantly hinder model performance.\\

\begin{note}[label={note:4}]{The performance of FAIREDU and state-of-the-art methods}
FAIREDU demonstrated no significant performance trade-off compared to the original model and other augmentation methods.
\end{note}

\begin{comment}
    
\begin{figure}[htbp]
    \centering
    \begin{subfigure}[b]{0.3\textwidth}
        \includegraphics[width=\textwidth]{Figure/Fig2.png}
        \caption{\(|1-DI|\) in Logistic Regression}
        \label{Fig.2}
    \end{subfigure}
    \hfill
    \begin{subfigure}[b]{0.3\textwidth}
        \includegraphics[width=\textwidth]{Figure/Fig3.png}
        \caption{\(|1-DI|\) in Random Forest}
        \label{Fig.3}
    \end{subfigure}
    \hfill
    \begin{subfigure}[b]{0.3\textwidth}
        \includegraphics[width=\textwidth]{Figure/Fig4.png}
        \caption{\(|1-DI|\) in Decision Tree}
        \label{Fig.4}
    \end{subfigure}
    \caption{Comparison of \(|1-DI|\) across different models}
    \label{fig:comparison1}
\end{figure}
\end{comment}

\section{Discussion}
\subsection{Answering RQs}
This section summarizes how the results presented in Section \ref{sec: Results} address each of the research questions posed in this paper. The findings provide valuable insights into the evaluation of fairness across multiple dimensions within educational datasets and demonstrate the effectiveness of FAIREDU as a fairness intervention. Below is a detailed breakdown of the answers to each research question:
\begin{itemize}
    \item Regarding RQ1 (addressing in Subsection \ref{subsec: RQ1}) , the results confirm the absence of significant bias among sensitive features in the educational datasets. Despite the presence of features like disability, health status, debtor status, and birthplace in only a single dataset, features such as gender, race, and age do not exhibit consistent bias across datasets. This emphasizes the need to assess all sensitive features for their potential impact on fairness and highlights the importance of developing interventions that can address multiple sensitive features simultaneously within a single dataset.
    \item Regarding RQ2 (addressing in Subsection \ref{subsec: RQ2}) show that different machine learning models yield varying fairness evaluations, even when applied to the same dataset and fairness metric. Decision Tree and Random Forest models, for example, demonstrate higher fairness compared to Logistic Regression models. Additionally, different fairness metrics produce different outcomes across models, highlighting the necessity of selecting appropriate fairness indices for each machine learning model to ensure accurate fairness assessments.
    \item Regarding RQ3 (addressing in Subsection \ref{subsec: RQ3}) demonstrates that FAIREDU effectively reduces the dependence of most features on sensitive features, enhancing overall fairness. Moreover, FAIREDU outperforms most current methods in terms of fairness, especially for multiple sensitive features. These results indicate that FAIREDU successfully tackles two major challenges: improving fairness and addressing fairness issues for multiple sensitive features simultaneously.
    \item Finally, regarding RQ4 (addressing in Subsection \ref{subsec: RQ4}), the results show that FAIREDU maintains strong performance in terms of accuracy (ACC) and recall, with minimal trade-offs. In many cases, the model's performance even improves. While this is a promising outcome, particularly given the ongoing challenges in balancing fairness and performance, further exploration with additional datasets and models is required to validate these findings.
\end{itemize}
%In summary, the results presented in this section provide comprehensive answers to the research questions outlined in this paper. The analysis confirms the absence of significant bias among the sensitive features within the educational datasets and underscores the importance of assessing all sensitive attributes to ensure fairness. Furthermore, the findings highlight the variability in fairness outcomes across different machine learning models and fairness metrics, reinforcing the need for careful model and metric selection in fairness evaluations.\\

Most notably, FAIREDU has proven to be an effective intervention, addressing fairness challenges across multiple sensitive features simultaneously while maintaining, and in some cases improving the model's performance. These results demonstrate that FAIREDU holds great potential as a robust tool for ensuring fairness in machine-learning applications, especially in complex datasets with multiple sensitive features. 

\subsection{Limitations}
While FAIREDU demonstrates promising capabilities in enhancing fairness across multiple sensitive features within educational datasets, several limitations concerning internal validity, external validity, construct validity, and conclusion validity must be acknowledged \citep{Yin_CaseStudy_2003,Runeson_2009,Cruzes_Dyba_2011}. To ensure the validity of this study, we adhered to the validity guidelines from Runeson \citep{Runeson_2009}.

\subsubsection{Internal Validity}
FAIREDU relies on multivariate linear regression to detect and eliminate dependencies between features and sensitive features. This linear assumption may limit the method’s ability to capture non-linear relationships inherent in certain datasets, potentially leaving some residual biases unaddressed. In a relevant work by Li et al. \citep{b23}, the authors compare the results of the linear regression and polynomial regression, showing a significantly better performance of linear regression than that of polynomial regressions. 

Besides, our evaluation focused on specific fairness metrics, and while these are widely recognized, they may not encompass all fairness dimensions relevant to every educational context. The selection of these metrics could influence the outcomes, potentially overlooking other significant aspects of fairness.

\subsubsection{External Validity}
The evaluation of FAIREDU was conducted using datasets specific to the education sector. Although chosen to represent various educational contexts, these datasets may not fully capture the diversity of real-world educational environments, leaving the effectiveness of FAIREDU in more diverse settings uncertain. Since our study relies on traditional ML algorithms (LR, RF, RT), the generalizability of our findings to more modern ML/AI approaches, such as Neural Networks, Deep Learning, etc, is limited.

\subsubsection{Construct Validity}
While FAIREDU addresses multiple sensitive features, the complex interactions between various intersectional identities may present challenges that the current model does not fully capture. Theoretically, FAIREDU can be applied to both discrete and continuous variables. However, in the education sector, it is common for sensitive features to be discrete and for outcome features to be binary. This may introduce limitations in the model’s ability to fully address fairness in these contexts.

\subsubsection{Reliability}
FAIREDU aims to balance fairness and predictive performance; however, trade-offs may still exist, particularly in cases involving highly imbalanced or conflicting sensitive features. Improving fairness for one set of features could unintentionally affect the performance or fairness of others, despite FAIREDU’s robust handling of multiple features. Moreover, modifying the dataset to remove dependencies may inadvertently alter other important relationships within the data, potentially impacting the interpretability and utility of the resulting machine-learning models. The majority of our experiments used datasets containing only two or three sensitive features. In the future, we plan to extend this work to a broader range of datasets with a greater number of sensitive features, in order to further validate the robustness and accuracy of the proposed method.

\section{Conclusion}
In this paper, we propose a method called FAIREDU to improve the fairness in the preprocessing of machine learning models focusing on the education domain. This method has shown its superiority when simultaneously solving significant problems in fairness research in machine learning, which are (1) providing a solution to improve the fairness for multiple sensitive features at the same time for datasets containing many sensitive features, (2) FAIREDU shows its superiority in terms of fairness improvement compared to previous basic and state-of-the-art methods such as Reweiging, Dir, Fairway, Fair-Smote, LTDD, (3) FAIREDU also shows its superiority when it shows that it also limits the possibility of model performance trade-offs after fairness intervention. All of these show that FAIREDU is indeed an effective method as it improves the major problems in current fairness research. These results show that FAIREDU is a promising tool for ensuring fairness in machine learning applications, especially in complex datasets with many sensitive features. In addition, our study also evaluates the impact of sensitive features and machine learning models on fairness.

Our study also identifies areas for further exploration. One important direction for future research is the expansion of datasets and models. To validate the generalizability of FAIREDU, it is essential to test its effectiveness across a broader range of datasets and machine learning models, including those from diverse domains with varying complexities and characteristics. Besides, future work should explore FAIREDU's performance with datasets that contain a wider array of sensitive features, particularly those that are less commonly studied, to assess its ability to address a broad spectrum of fairness challenges. Another promising area of future research is the enhancement of fairness methods and metrics. This includes the development of composite sensitive features derived from existing ones within a dataset and providing a general solution for improving fairness across different datasets. Furthermore, researchers should explore the creation of new fairness metrics that offer more nuanced evaluations, especially in scenarios involving multiple sensitive features. Conducting sensitivity analyses on various fairness metrics will also be crucial to understanding how these metrics impact model performance and fairness, and to identify the most appropriate metrics for different applications. Lastly, future research should focus on the trade-offs between model performance and fairness. Investigating methods to balance fairness and performance effectively is critical, particularly in developing innovative methods that minimize performance trade-offs while enhancing fairness. Additionally, the development of adaptive fairness interventions that can dynamically adjust based on the model’s performance and fairness needs will be an essential step in ensuring optimal outcomes in diverse machine learning scenarios. Addressing these areas will advance the field of fair machine learning and contribute to the development of more equitable and effective AI systems.

\bibliographystyle{elsarticle-num}
\bibliography{main}

\begin{thebibliography}{10}
\expandafter\ifx\csname url\endcsname\relax
  \def\url#1{\texttt{#1}}\fi
\expandafter\ifx\csname urlprefix\endcsname\relax\def\urlprefix{URL }\fi
\expandafter\ifx\csname href\endcsname\relax
  \def\href#1#2{#2} \def\path#1{#1}\fi

\bibitem{nguyen-duc_generative_2024}
A.~Nguyen-Duc, P.~Abrahamsson, F.~Khomh (Eds.), \href{https://link.springer.com/10.1007/978-3-031-55642-5}{Generative {AI} for {Effective} {Software} {Development}}, Springer Nature Switzerland, Cham, 2024.
\newblock \href {https://doi.org/10.1007/978-3-031-55642-5} {\path{doi:10.1007/978-3-031-55642-5}}.
\newline\urlprefix\url{https://link.springer.com/10.1007/978-3-031-55642-5}

\bibitem{peng_fairmask_2023}
K.~Peng, J.~Chakraborty, T.~Menzies, {FairMask}: {Better} {Fairness} via {Model}-{Based} {Rebalancing} of {Protected} {Attributes}, IEEE Trans. Softw. Eng. 49~(4) (2023) 2426--2439.
\newblock \href {https://doi.org/10.1109/TSE.2022.3220713} {\path{doi:10.1109/TSE.2022.3220713}}.

\bibitem{b1}
Z.~Chen, J.~M. Zhang, M.~Hort, M.~Harman, F.~Sarro, \href{https://doi.org/10.1145/3652155}{Fairness testing: A comprehensive survey and analysis of trends}, ACM Trans. Softw. Eng. Methodol. 33~(5) (Jun. 2024).
\newblock \href {https://doi.org/10.1145/3652155} {\path{doi:10.1145/3652155}}.
\newline\urlprefix\url{https://doi.org/10.1145/3652155}

\bibitem{b2}
S.~Biswas, H.~Rajan, \href{https://doi.org/10.1145/3368089.3409704}{Do the {Machine} {Learning} {Models} on a {Crowd} {Sourced} {Platform} {Exhibit} {Bias}? {An} {Empirical} {Study} on {Model} {Fairness}}, in: Proceedings of the 28th {ACM} {Joint} {Meeting} on {European} {Software} {Engineering} {Conference} and {Symposium} on the {Foundations} of {Software} {Engineering}, 2020, pp. 642--653, arXiv:2005.12379 [cs, stat].
\newblock \href {https://doi.org/10.1145/3368089.3409704} {\path{doi:10.1145/3368089.3409704}}.
\newline\urlprefix\url{https://doi.org/10.1145/3368089.3409704}

\bibitem{b3}
S.~Biswas, H.~Rajan, \href{https://doi.org/10.1145/3468264.3468536}{Fair {Preprocessing}: {Towards} {Understanding} {Compositional} {Fairness} of {Data} {Transformers} in {Machine} {Learning} {Pipeline}}, in: Proceedings of the 29th {ACM} {Joint} {Meeting} on {European} {Software} {Engineering} {Conference} and {Symposium} on the {Foundations} of {Software} {Engineering}, 2021, pp. 981--993, arXiv:2106.06054 [cs].
\newblock \href {https://doi.org/10.1145/3468264.3468536} {\path{doi:10.1145/3468264.3468536}}.
\newline\urlprefix\url{https://doi.org/10.1145/3468264.3468536}

\bibitem{b4}
J.~Chakraborty, S.~Majumder, T.~Menzies, \href{https://doi.org/10.1145/3468264.3468537}{Bias in {Machine} {Learning} {Software}: {Why}? {How}? {What} to do?}, in: Proceedings of the 29th {ACM} {Joint} {Meeting} on {European} {Software} {Engineering} {Conference} and {Symposium} on the {Foundations} of {Software} {Engineering}, 2021, pp. 429--440.
\newblock \href {https://doi.org/10.1145/3468264.3468537} {\path{doi:10.1145/3468264.3468537}}.
\newline\urlprefix\url{https://doi.org/10.1145/3468264.3468537}

\bibitem{b5}
J.~Chakraborty, S.~Majumder, Z.~Yu, T.~Menzies, \href{https://doi.org/10.1145/3368089.3409697}{Fairway: {A} {Way} to {Build} {Fair} {ML} {Software}}, in: Proceedings of the 28th {ACM} {Joint} {Meeting} on {European} {Software} {Engineering} {Conference} and {Symposium} on the {Foundations} of {Software} {Engineering}, 2020, pp. 654--665.
\newblock \href {https://doi.org/10.1145/3368089.3409697} {\path{doi:10.1145/3368089.3409697}}.
\newline\urlprefix\url{https://doi.org/10.1145/3368089.3409697}

\bibitem{b6}
U.~Gohar, S.~Biswas, H.~Rajan, \href{https://ieeexplore.ieee.org/document/10172501}{Towards {Understanding} {Fairness} and its {Composition} in {Ensemble} {Machine} {Learning}}, in: 2023 {IEEE}/{ACM} 45th {International} {Conference} on {Software} {Engineering} ({ICSE}), 2023, pp. 1533--1545.
\newblock \href {https://doi.org/10.1109/ICSE48619.2023.00133} {\path{doi:10.1109/ICSE48619.2023.00133}}.
\newline\urlprefix\url{https://ieeexplore.ieee.org/document/10172501}

\bibitem{b7}
M.~Hort, F.~Sarro, \href{https://ieeexplore.ieee.org/document/9678568/}{Did {You} {Do} {Your} {Homework}? {Raising} {Awareness} on {Software} {Fairness} and {Discrimination}}, 2021 36th IEEE/ACM International Conference on Automated Software Engineering (ASE) (2021) 1322--1326Conference Name: 2021 36th IEEE/ACM International Conference on Automated Software Engineering (ASE) ISBN: 9781665403375 Place: Melbourne, Australia Publisher: IEEE.
\newblock \href {https://doi.org/10.1109/ASE51524.2021.9678568} {\path{doi:10.1109/ASE51524.2021.9678568}}.
\newline\urlprefix\url{https://ieeexplore.ieee.org/document/9678568/}

\bibitem{b8}
M.~Hort, J.~M. Zhang, F.~Sarro, M.~Harman, \href{https://doi.org/10.1145/3468264.3468565}{Fairea: a model behaviour mutation approach to benchmarking bias mitigation methods}, in: Proceedings of the 29th {ACM} {Joint} {Meeting} on {European} {Software} {Engineering} {Conference} and {Symposium} on the {Foundations} of {Software} {Engineering}, {ESEC}/{FSE} 2021, Association for Computing Machinery, New York, NY, USA, 2021, pp. 994--1006.
\newblock \href {https://doi.org/10.1145/3468264.3468565} {\path{doi:10.1145/3468264.3468565}}.
\newline\urlprefix\url{https://doi.org/10.1145/3468264.3468565}

\bibitem{b9}
J.~M. Zhang, M.~Harman, \href{https://doi.org/10.1109/ICSE43902.2021.00129}{"{Ignorance} and {Prejudice}" in {Software} {Fairness}}, in: Proceedings of the 43rd {International} {Conference} on {Software} {Engineering}, {ICSE} '21, IEEE Press, Madrid, Spain, 2021, pp. 1436--1447.
\newblock \href {https://doi.org/10.1109/ICSE43902.2021.00129} {\path{doi:10.1109/ICSE43902.2021.00129}}.
\newline\urlprefix\url{https://doi.org/10.1109/ICSE43902.2021.00129}

\bibitem{b11}
L.~Hardesty, \href{https://news.mit.edu/2018/study-finds-gender-skin-type-bias-artificial-intelligence-systems-0212}{Study finds gender and skin-type bias in commercial artificial-intelligence systems {\textbar} {MIT} {News} {\textbar} {Massachusetts} {Institute} of {Technology}}, accessed date: 24 May 2024.
\newline\urlprefix\url{https://news.mit.edu/2018/study-finds-gender-skin-type-bias-artificial-intelligence-systems-0212}

\bibitem{b12}
\href{https://www.eeoc.gov/initiatives/e-race/significant-eeoc-racecolor-casescovering-private-and-federal-sectors#intersectional}{Significant {EEOC} {Race}/{Color} {Cases}({Covering} {Private} and {Federal} {Sectors}) {\textbar} {U}.{S}. {Equal} {Employment} {Opportunity} {Commission}}.
\newline\urlprefix\url{https://www.eeoc.gov/initiatives/e-race/significant-eeoc-racecolor-casescovering-private-and-federal-sectors#intersectional}

\bibitem{b13}
X.~Li, Z.~Chen, J.~Zhang, F.~Sarro, Y.~Zhang, X.~Liu, Dark-{Skin} {Individuals} {Are} at {More} {Risk} on the {Street}: {Unmasking} {Fairness} {Issues} of {Autonomous} {Driving} {Systems}, 2023.

\bibitem{pham_fairness_2024}
N.~Pham, A.~Nguyen-Duc, H.~Pham-Ngoc, \href{https://doi.org/10.2139/ssrn.4713827}{Fairness for machine learning software in education: A systematic mapping study}, The Journal of System and Software (oct 2024).
\newline\urlprefix\url{https://doi.org/10.2139/ssrn.4713827}

\bibitem{minnaert_1997}
A.~Minnaert, P.~J. Janssen, Bias in the assessment of regulation activities in studying at the level of higher education, Eur. J. Psychol. Assess. 13~(2) (1997) 99--108.
\newblock \href {https://doi.org/10.1027/1015-5759.13.2.99} {\path{doi:10.1027/1015-5759.13.2.99}}.

\bibitem{engberg_2004}
M.~E. Engberg, Improving intergroup relations in higher education: A critical examination of the influence of educational interventions on racial bias, Rev. Educ. Res. 74~(4) (2004) 473--524.
\newblock \href {https://doi.org/10.3102/00346543074004473} {\path{doi:10.3102/00346543074004473}}.

\bibitem{huston_2006}
T.~Huston, \href{https://digitalcommons.law.seattleu.edu/sjsj/vol4/iss2/34}{Race and gender bias in higher education: Could faculty course evaluations impede further progress toward parity?}, Seattle J. Soc. Justice 4~(2) (May 2006).
\newline\urlprefix\url{https://digitalcommons.law.seattleu.edu/sjsj/vol4/iss2/34}

\bibitem{hughes_2013}
G.~Hughes, Racial justice, hegemony, and bias incidents in u.s. higher education, Multicultural Perspectives (2013) 126–132\href {https://doi.org/10.1080/15210960.2013.809301} {\path{doi:10.1080/15210960.2013.809301}}.

\bibitem{mahmud_2020}
A.~Mahmud, Racial disparities in student outcomes in british higher education: Examining mindsets and bias, Teaching in Higher Education (2020) 254–269\href {https://doi.org/10.1080/13562517.2020.1796619} {\path{doi:10.1080/13562517.2020.1796619}}.

\bibitem{pessach_2022}
D.~Pessach, E.~Shmueli, \href{https://doi.org/10.1145/3494672}{A review on fairness in machine learning}, ACM Comput. Surv. 55~(3) (2022) 51:1--51:44.
\newblock \href {https://doi.org/10.1145/3494672} {\path{doi:10.1145/3494672}}.
\newline\urlprefix\url{https://doi.org/10.1145/3494672}

\bibitem{zhai_2020}
X.~Zhai, et~al., A review of artificial intelligence (ai) in education from 2010 to 2020, Complexity 2021 (2021) 1--18.
\newblock \href {https://doi.org/10.1155/2021/8812542} {\path{doi:10.1155/2021/8812542}}.

\bibitem{b28}
J.~R. Foulds, R.~Islam, K.~N. Keya, S.~Pan, \href{https://ieeexplore.ieee.org/document/9101635/}{An {Intersectional} {Definition} of {Fairness}}, 2020 IEEE 36th International Conference on Data Engineering (ICDE) (2020) 1918--1921Conference Name: 2020 IEEE 36th International Conference on Data Engineering (ICDE) ISBN: 9781728129037 Place: Dallas, TX, USA Publisher: IEEE.
\newblock \href {https://doi.org/10.1109/ICDE48307.2020.00203} {\path{doi:10.1109/ICDE48307.2020.00203}}.
\newline\urlprefix\url{https://ieeexplore.ieee.org/document/9101635/}

\bibitem{b29}
K.~Crenshaw, \href{https://doi.org/10.1093/oso/9780198782063.003.0016}{Demarginalizing the {Intersection} of {Race} and {Sex}: {A} {Black} {Feminist} {Critique} of {Antidiscrimination} {Doctrine}, {Feminist} {Theory}, and {Antiracist} {Politics}}, in: A.~Phillips (Ed.), Feminism {And} {Politics}: {Oxford} {Readings} {In} {Feminism}, Oxford University Press, 1998, p.~0.
\newblock \href {https://doi.org/10.1093/oso/9780198782063.003.0016} {\path{doi:10.1093/oso/9780198782063.003.0016}}.
\newline\urlprefix\url{https://doi.org/10.1093/oso/9780198782063.003.0016}

\bibitem{b32}
Z.~Chen, J.~M. Zhang, F.~Sarro, M.~Harman, \href{https://doi.org/10.1145/3583561}{A comprehensive empirical study of bias mitigation methods for machine learning classifiers}, Vol.~32, Association for Computing Machinery, New York, NY, USA, 2023.
\newblock \href {https://doi.org/10.1145/3583561} {\path{doi:10.1145/3583561}}.
\newline\urlprefix\url{https://doi.org/10.1145/3583561}

\bibitem{b33}
R.~Berk, H.~Heidari, S.~Jabbari, M.~Kearns, A.~Roth, \href{http://journals.sagepub.com/doi/10.1177/0049124118782533}{Fairness in {Criminal} {Justice} {Risk} {Assessments}: {The} {State} of the {Art}}, Sociological Methods \& Research 50~(1) (2021) 3--44.
\newblock \href {https://doi.org/10.1177/0049124118782533} {\path{doi:10.1177/0049124118782533}}.
\newline\urlprefix\url{http://journals.sagepub.com/doi/10.1177/0049124118782533}

\bibitem{b34}
S.~Corbett-Davies, E.~Pierson, A.~Feller, S.~Goel, A.~Huq, \href{https://doi.org/10.1145/3097983.3098095}{Algorithmic {Decision} {Making} and the {Cost} of {Fairness}}, in: Proceedings of the 23rd {ACM} {SIGKDD} {International} {Conference} on {Knowledge} {Discovery} and {Data} {Mining}, {KDD} '17, Association for Computing Machinery, New York, NY, USA, 2017, pp. 797--806.
\newblock \href {https://doi.org/10.1145/3097983.3098095} {\path{doi:10.1145/3097983.3098095}}.
\newline\urlprefix\url{https://doi.org/10.1145/3097983.3098095}

\bibitem{b35}
M.~Wick, S.~Panda, J.-B. Tristan, Unlocking fairness: a trade-off revisited, in: Proceedings of the 33rd International Conference on Neural Information Processing Systems, Curran Associates Inc., Red Hook, NY, USA, 2019.

\bibitem{b16}
N.~Pham, H.~Pham-Ngoc, A.~Nguyen-Duc, Fairness {Requirement} in {AI} {Engineering} – {A} {Review} on {Current} {Research} and {Future} {Directions}, in: V.~Gupta, L.~Rubalcaba, C.~Gupta, T.~Hanne (Eds.), Sustainability in {Software} {Engineering} and {Business} {Information} {Management}, Springer International Publishing, Cham, 2023, pp. 3--13.
\newblock \href {https://doi.org/10.1007/978-3-031-32436-9_1} {\path{doi:10.1007/978-3-031-32436-9_1}}.

\bibitem{b17}
Z.~Chen, J.~M. Zhang, F.~Sarro, M.~Harman, \href{https://doi.org/10.1145/3597503.3639083}{Fairness improvement with multiple protected attributes: How far are we?}, in: Proceedings of the IEEE/ACM 46th International Conference on Software Engineering, ICSE '24, Association for Computing Machinery, New York, NY, USA, 2024.
\newblock \href {https://doi.org/10.1145/3597503.3639083} {\path{doi:10.1145/3597503.3639083}}.
\newline\urlprefix\url{https://doi.org/10.1145/3597503.3639083}

\bibitem{b18}
F.~Kamiran, T.~Calders, \href{https://doi.org/10.1007/s10115-011-0463-8}{Data preprocessing techniques for classification without discrimination}, Knowledge and Information Systems 33~(1) (2012) 1--33.
\newblock \href {https://doi.org/10.1007/s10115-011-0463-8} {\path{doi:10.1007/s10115-011-0463-8}}.
\newline\urlprefix\url{https://doi.org/10.1007/s10115-011-0463-8}

\bibitem{b19}
M.~Feldman, S.~A. Friedler, J.~Moeller, C.~Scheidegger, S.~Venkatasubramanian, \href{https://doi.org/10.1145/2783258.2783311}{Certifying and removing disparate impact}, in: Proceedings of the 21th ACM SIGKDD International Conference on Knowledge Discovery and Data Mining, KDD '15, Association for Computing Machinery, New York, NY, USA, 2015, p. 259–268.
\newblock \href {https://doi.org/10.1145/2783258.2783311} {\path{doi:10.1145/2783258.2783311}}.
\newline\urlprefix\url{https://doi.org/10.1145/2783258.2783311}

\bibitem{b20}
L.~E. Celis, L.~Huang, V.~Keswani, N.~K. Vishnoi, \href{https://doi.org/10.1145/3287560.3287586}{Classification with fairness constraints: A meta-algorithm with provable guarantees}, in: Proceedings of the Conference on Fairness, Accountability, and Transparency, FAT* '19, Association for Computing Machinery, New York, NY, USA, 2019, p. 319–328.
\newblock \href {https://doi.org/10.1145/3287560.3287586} {\path{doi:10.1145/3287560.3287586}}.
\newline\urlprefix\url{https://doi.org/10.1145/3287560.3287586}

\bibitem{Mitigating_Unwanted_Biases}
B.~H. Zhang, B.~Lemoine, M.~Mitchell, \href{https://doi.org/10.1145/3278721.3278779}{Mitigating unwanted biases with adversarial learning}, in: Proceedings of the 2018 AAAI/ACM Conference on AI, Ethics, and Society, AIES '18, Association for Computing Machinery, New York, NY, USA, 2018, p. 335–340.
\newblock \href {https://doi.org/10.1145/3278721.3278779} {\path{doi:10.1145/3278721.3278779}}.
\newline\urlprefix\url{https://doi.org/10.1145/3278721.3278779}

\bibitem{b22}
T.~Kamishima, S.~Akaho, H.~Asoh, J.~Sakuma, Fairness-{Aware} {Classifier} with {Prejudice} {Remover} {Regularizer}, in: P.~A. Flach, T.~De~Bie, N.~Cristianini (Eds.), Machine {Learning} and {Knowledge} {Discovery} in {Databases}, Springer, Berlin, Heidelberg, 2012, pp. 35--50.
\newblock \href {https://doi.org/10.1007/978-3-642-33486-3_3} {\path{doi:10.1007/978-3-642-33486-3_3}}.

\bibitem{b23}
M.~Hardt, E.~Price, N.~Srebro, Equality of opportunity in supervised learning, in: Proceedings of the 30th International Conference on Neural Information Processing Systems, NIPS'16, Curran Associates Inc., Red Hook, NY, USA, 2016, p. 3323–3331.

\bibitem{b24}
G.~Pleiss, M.~Raghavan, F.~Wu, J.~Kleinberg, K.~Q. Weinberger, On fairness and calibration, in: Proceedings of the 31st International Conference on Neural Information Processing Systems, NIPS'17, Curran Associates Inc., Red Hook, NY, USA, 2017, p. 5684–5693.

\bibitem{b25}
F.~Kamiran, A.~Karim, X.~Zhang, \href{https://ieeexplore.ieee.org/document/6413831}{Decision {Theory} for {Discrimination}-{Aware} {Classification}}, in: 2012 {IEEE} 12th {International} {Conference} on {Data} {Mining}, 2012, pp. 924--929, iSSN: 2374-8486.
\newblock \href {https://doi.org/10.1109/ICDM.2012.45} {\path{doi:10.1109/ICDM.2012.45}}.
\newline\urlprefix\url{https://ieeexplore.ieee.org/document/6413831}

\bibitem{b15}
Z.~Chen, J.~M. Zhang, F.~Sarro, M.~Harman, \href{https://doi.org/10.1145/3540250.3549093}{{MAAT}: a novel ensemble approach to addressing fairness and performance bugs for machine learning software}, in: Proceedings of the 30th {ACM} {Joint} {European} {Software} {Engineering} {Conference} and {Symposium} on the {Foundations} of {Software} {Engineering}, {ESEC}/{FSE} 2022, Association for Computing Machinery, New York, NY, USA, 2022, pp. 1122--1134.
\newblock \href {https://doi.org/10.1145/3540250.3549093} {\path{doi:10.1145/3540250.3549093}}.
\newline\urlprefix\url{https://doi.org/10.1145/3540250.3549093}

\bibitem{b26}
K.~Peng, J.~Chakraborty, T.~Menzies, \href{https://ieeexplore.ieee.org/document/9951398}{{FairMask}: {Better} {Fairness} via {Model}-{Based} {Rebalancing} of {Protected} {Attributes}}, IEEE Transactions on Software Engineering 49~(4) (2023) 2426--2439, conference Name: IEEE Transactions on Software Engineering.
\newblock \href {https://doi.org/10.1109/TSE.2022.3220713} {\path{doi:10.1109/TSE.2022.3220713}}.
\newline\urlprefix\url{https://ieeexplore.ieee.org/document/9951398}

\bibitem{b27}
Y.~Li, L.~Meng, L.~Chen, L.~Yu, D.~Wu, Y.~Zhou, B.~Xu, \href{https://ieeexplore.ieee.org/document/9794106}{Training {Data} {Debugging} for the {Fairness} of {Machine} {Learning} {Software}}, in: 2022 {IEEE}/{ACM} 44th {International} {Conference} on {Software} {Engineering} ({ICSE}), 2022, pp. 2215--2227, iSSN: 1558-1225.
\newblock \href {https://doi.org/10.1145/3510003.3510091} {\path{doi:10.1145/3510003.3510091}}.
\newline\urlprefix\url{https://ieeexplore.ieee.org/document/9794106}

\bibitem{b36}
N.~A. Saxena, K.~Huang, E.~DeFilippis, G.~Radanovic, D.~C. Parkes, Y.~Liu, \href{https://dl.acm.org/doi/10.1145/3306618.3314248}{How {Do} {Fairness} {Definitions} {Fare}?: {Examining} {Public} {Attitudes} {Towards} {Algorithmic} {Definitions} of {Fairness}}, in: Proceedings of the 2019 {AAAI}/{ACM} {Conference} on {AI}, {Ethics}, and {Society}, ACM, Honolulu HI USA, 2019, pp. 99--106.
\newblock \href {https://doi.org/10.1145/3306618.3314248} {\path{doi:10.1145/3306618.3314248}}.
\newline\urlprefix\url{https://dl.acm.org/doi/10.1145/3306618.3314248}

\bibitem{b10}
N.~Mehrabi, F.~Morstatter, N.~Saxena, K.~Lerman, A.~Galstyan, A survey on bias and fairness in machine learning, ACM Comput. Surv. 54~(6) (Jul. 2021).
\newblock \href {https://doi.org/10.1145/3457607} {\path{doi:10.1145/3457607}}.

\bibitem{b37}
K.~Zhang, A.~Aslan, Ai technologies for education: Recent research and future directions, Comput. Educ. Artif. Intell. 2 (2021) 100025.
\newblock \href {https://doi.org/10.1016/j.caeai.2021.100025} {\path{doi:10.1016/j.caeai.2021.100025}}.

\bibitem{b39}
S.~Verma, J.~Rubin, Fairness definitions explained, in: Proceedings of the International Workshop on Software Fairness, FairWare '18, Association for Computing Machinery, New York, NY, USA, 2018, pp. 1--7.
\newblock \href {https://doi.org/10.1145/3194770.3194776} {\path{doi:10.1145/3194770.3194776}}.

\bibitem{b40}
M.~Kusner, J.~Loftus, C.~Russell, R.~Silva, Counterfactual fairness, in: Proceedings of the 31st International Conference on Neural Information Processing Systems, NIPS'17, Curran Associates Inc., Red Hook, NY, USA, 2017, p. 4069–4079.

\bibitem{b41}
C.~Dwork, M.~Hardt, T.~Pitassi, O.~Reingold, R.~Zemel, \href{https://doi.org/10.1145/2090236.2090255}{Fairness through awareness}, in: Proceedings of the 3rd Innovations in Theoretical Computer Science Conference, ITCS '12, Association for Computing Machinery, New York, NY, USA, 2012, p. 214–226.
\newblock \href {https://doi.org/10.1145/2090236.2090255} {\path{doi:10.1145/2090236.2090255}}.
\newline\urlprefix\url{https://doi.org/10.1145/2090236.2090255}

\bibitem{b43}
R.~Berk, H.~Heidari, S.~Jabbari, M.~Kearns, A.~Roth, Fairness in criminal justice risk assessments: The state of the art, Sociol. Methods Res. 50~(1) (2021) 3--44.
\newblock \href {https://doi.org/10.1177/0049124118782533} {\path{doi:10.1177/0049124118782533}}.

\bibitem{b44}
A.~Chouldechova, Fair prediction with disparate impact: A study of bias in recidivism prediction instruments, Big data 5 2 (2016) 153--163.

\bibitem{b45}
G.~Farnadi, B.~Babaki, L.~Getoor, Fairness in relational domains, in: Proceedings of the 2018 AAAI/ACM Conference on AI, Ethics, and Society, ACM, New Orleans LA USA, 2018, pp. 108--114.
\newblock \href {https://doi.org/10.1145/3278721.3278733} {\path{doi:10.1145/3278721.3278733}}.

\bibitem{b47}
S.~Biswas, H.~Rajan, \href{https://doi.org/10.1109/ICSE48619.2023.00134}{Fairify: Fairness verification of neural networks}, in: ICSE'23: The 45th International Conference on Software Engineering, 2023, p. 1546–1558.
\newblock \href {https://doi.org/10.1109/ICSE48619.2023.00134} {\path{doi:10.1109/ICSE48619.2023.00134}}.
\newline\urlprefix\url{https://doi.org/10.1109/ICSE48619.2023.00134}

\bibitem{sweeney_discrimination_2013}
L.~Sweeney, \href{https://dl.acm.org/doi/10.1145/2447976.2447990}{Discrimination in online ad delivery}, Commun. ACM 56~(5) (2013) 44--54.
\newblock \href {https://doi.org/10.1145/2447976.2447990} {\path{doi:10.1145/2447976.2447990}}.
\newline\urlprefix\url{https://dl.acm.org/doi/10.1145/2447976.2447990}

\bibitem{ribeiro_anchors_2018}
M.~T. Ribeiro, S.~Singh, C.~Guestrin, \href{https://ojs.aaai.org/index.php/AAAI/article/view/11491}{Anchors: {High}-{Precision} {Model}-{Agnostic} {Explanations}}, Proceedings of the AAAI Conference on Artificial Intelligence 32~(1), number: 1 (Apr. 2018).
\newblock \href {https://doi.org/10.1609/aaai.v32i1.11491} {\path{doi:10.1609/aaai.v32i1.11491}}.
\newline\urlprefix\url{https://ojs.aaai.org/index.php/AAAI/article/view/11491}

\bibitem{barry_becker_adult_1996}
R.~K. Barry~Becker, \href{https://archive.ics.uci.edu/dataset/2}{Adult}, accessed date: 27 August 2024 (1996).
\newblock \href {https://doi.org/10.24432/C5XW20} {\path{doi:10.24432/C5XW20}}.
\newline\urlprefix\url{https://archive.ics.uci.edu/dataset/2}

\bibitem{noauthor_propublicacompas-analysis_2024}
\href{https://github.com/propublica/compas-analysis}{Compas analysis}, accessed date: 27 August 2024 (Aug. 2024).
\newline\urlprefix\url{https://github.com/propublica/compas-analysis}

\bibitem{i-cheng_yeh_default_2009}
{I-Cheng Yeh}, \href{https://archive.ics.uci.edu/dataset/350}{Default of {Credit} {Card} {Clients}}, accessed date: 27 August 2024 (2009).
\newblock \href {https://doi.org/10.24432/C55S3H} {\path{doi:10.24432/C55S3H}}.
\newline\urlprefix\url{https://archive.ics.uci.edu/dataset/350}

\bibitem{noauthor_predict_nodate_dropout}
\href{https://www.kaggle.com/datasets/thedevastator/higher-education-predictors-of-student-retention}{Predict students' dropout and academic success}, accessed date: 07 July 2024.
\newline\urlprefix\url{https://www.kaggle.com/datasets/thedevastator/higher-education-predictors-of-student-retention}

\bibitem{cortez_student_Performance_2008}
P.~Cortez, \href{https://archive.ics.uci.edu/dataset/320}{Student {Performance}} (2008).
\newblock \href {https://doi.org/10.24432/C5TG7T} {\path{doi:10.24432/C5TG7T}}.
\newline\urlprefix\url{https://archive.ics.uci.edu/dataset/320}

\bibitem{kuzilek_open_2017_Oulad}
J.~Kuzilek, M.~Hlosta, Z.~Zdrahal, \href{https://www.nature.com/articles/sdata2017171}{Open {University} {Learning} {Analytics} dataset}, Scientific Data 4~(1) (2017) 170171.
\newblock \href {https://doi.org/10.1038/sdata.2017.171} {\path{doi:10.1038/sdata.2017.171}}.
\newline\urlprefix\url{https://www.nature.com/articles/sdata2017171}

\bibitem{kharb_role_2021}
L.~Kharb, P.~Singh, \href{https://www.igi-global.com/chapter/role-of-machine-learning-in-modern-education-and-teaching/www.igi-global.com/chapter/role-of-machine-learning-in-modern-education-and-teaching/261497}{Role of {Machine} {Learning} in {Modern} {Education} and {Teaching}}, in: Impact of {AI} {Technologies} on {Teaching}, {Learning}, and {Research} in {Higher} {Education}, IGI Global, 2021, pp. 99--123.
\newblock \href {https://doi.org/10.4018/978-1-7998-4763-2.ch006} {\path{doi:10.4018/978-1-7998-4763-2.ch006}}.
\newline\urlprefix\url{https://www.igi-global.com/chapter/role-of-machine-learning-in-modern-education-and-teaching/www.igi-global.com/chapter/role-of-machine-learning-in-modern-education-and-teaching/261497}

\bibitem{luan_review_2021}
H.~Luan, C.-C. Tsai, \href{https://www.jstor.org/stable/26977871}{A {Review} of {Using} {Machine} {Learning} {Approaches} for {Precision} {Education}}, Educational Technology \& Society 24~(1) (2021) 250--266, publisher: International Forum of Educational Technology \& Society.
\newline\urlprefix\url{https://www.jstor.org/stable/26977871}

\bibitem{A_review_2024}
S.~M.~L. ~, \href{https://www.ijfmr.com/research-paper.php?id=15481}{Review of {Machine} {Learning} {Models} for {Application} in {Adaptive} {Learning} for {Higher} {Education} {Student}}, International Journal For Multidisciplinary Research 6~(2) (2024) 15481.
\newblock \href {https://doi.org/10.36948/ijfmr.2024.v06i02.15481} {\path{doi:10.36948/ijfmr.2024.v06i02.15481}}.
\newline\urlprefix\url{https://www.ijfmr.com/research-paper.php?id=15481}

\bibitem{Costa2023}
V.~G. Costa, C.~E. Pedreira, \href{https://doi.org/10.1007/s10462-022-10275-5}{Recent advances in decision trees: an updated survey}, Artificial Intelligence Review 56 (2023) 4765–4800.
\newblock \href {https://doi.org/10.1007/s10462-022-10275-5} {\path{doi:10.1007/s10462-022-10275-5}}.
\newline\urlprefix\url{https://doi.org/10.1007/s10462-022-10275-5}

\bibitem{Random_Forest}
S.~Bernard, L.~Heutte, S.~Adam, On the selection of decision trees in random forests, in: 2009 International Joint Conference on Neural Networks, 2009, pp. 302--307.
\newblock \href {https://doi.org/10.1109/IJCNN.2009.5178693} {\path{doi:10.1109/IJCNN.2009.5178693}}.

\bibitem{acc_and_recall}
R.~Yacouby, D.~Axman, \href{https://aclanthology.org/2020.eval4nlp-1.9}{Probabilistic extension of precision, recall, and f1 score for more thorough evaluation of classification models}, in: S.~Eger, Y.~Gao, M.~Peyrard, W.~Zhao, E.~Hovy (Eds.), Proceedings of the First Workshop on Evaluation and Comparison of NLP Systems, Association for Computational Linguistics, Online, 2020, pp. 79--91.
\newblock \href {https://doi.org/10.18653/v1/2020.eval4nlp-1.9} {\path{doi:10.18653/v1/2020.eval4nlp-1.9}}.
\newline\urlprefix\url{https://aclanthology.org/2020.eval4nlp-1.9}

\bibitem{Yin_CaseStudy_2003}
T.~Hollweck, Robert k. yin. (2014). case study research design and methods (5th ed.). thousand oaks, ca: Sage. 282 pages., The Canadian Journal of Program Evaluation 30 (03 2016).
\newblock \href {https://doi.org/10.3138/cjpe.30.1.108} {\path{doi:10.3138/cjpe.30.1.108}}.

\bibitem{Runeson_2009}
P.~Runeson, M.~H{\"o}st, Guidelines for conducting and reporting case study research in software engineering, Empirical software engineering 14~(2) (2009) 131–164.
\newblock \href {https://doi.org/https://doi.org/10.1007/s10664-008-9102-8} {\path{doi:https://doi.org/10.1007/s10664-008-9102-8}}.

\bibitem{Cruzes_Dyba_2011}
D.~S. Cruzes, T.~Dyba, Recommended steps for thematic synthesis in software engineering, in: 2011 International Symposium on Empirical Software Engineering and Measurement, 2011, p. 275–284.
\newblock \href {https://doi.org/10.1109/ESEM.2011.36} {\path{doi:10.1109/ESEM.2011.36}}.

\end{thebibliography}

\end{document}